 \documentclass[sn-mathphys,Numbered]{sn-jnl}

\usepackage{graphicx}%
\usepackage{multirow}%
\usepackage{amsmath,amssymb,amsfonts}%
\usepackage{amsthm}%
\usepackage{mathrsfs}%
\usepackage[title]{appendix}%
\usepackage{xcolor}%
\usepackage{textcomp}%
\usepackage{manyfoot}%
\usepackage{booktabs}%
\usepackage{algorithm}%
\usepackage{algorithmicx}%
\usepackage{algpseudocode}%
\usepackage{listings}%
\usepackage{array} 
\usepackage{subcaption}%
\usepackage{adjustbox} 
\usepackage{soul}%

\begin{document}

\title{Unveiling the Human-like Similarities of Automatic Facial Expression Recognition: An Empirical Exploration through Explainable AI}


\author[1,2,3]{\fnm{F. Xavier} \sur{Gaya-Morey}}\email{francesc-xavier.gaya@uib.es}
\equalcont{These authors contributed equally to this work.}

\author[1,2,3]{\fnm{Silvia} \sur{Ramis-Guarinos}}\email{silvia.ramis@uib.es}
\equalcont{These authors contributed equally to this work.}

\author*[1,2,3]{\fnm{Cristina} \sur{Manresa-Yee}}\email{cristina.manresa@uib.es}
\equalcont{These authors contributed equally to this work.}

\author[1,2,3]{\fnm{José M.} \sur{Buades-Rubio}}\email{josemaria.buades@uib.es}\equalcont{These authors contributed equally to this work.}

\affil[1]{\orgdiv{Group of Computer Graphics, Computer Vision and AI (UGIVIA)}, \orgname{Universitat de les Illes Balears (UIB)}, \orgaddress{\street{Carretera de Valldemossa, km 7.5}, \city{Palma}, \postcode{07122}, \state{Illes Balears}, \country{Spain}}}

\affil[2]{\orgdiv{Research Institute of Health Sciences (IUNICS)}, \orgname{Universitat de les Illes Balears (UIB)}, \orgaddress{\street{Carretera de Valldemossa, km 7.5}, \city{Palma}, \postcode{07122}, \state{Illes Balears}, \country{Spain}}}

\affil[3]{\orgdiv{Department of Mathematics and Computer Science}, \orgname{Universitat de les Illes Balears (UIB)}, \orgaddress{\street{Carretera de Valldemossa, km 7.5}, \city{Palma}, \postcode{07122}, \state{Illes Balears}, \country{Spain}}}

\abstract{Facial expression recognition is vital for human behavior analysis, and deep learning has enabled models that can outperform humans. However, it is unclear how closely they mimic human processing. This study aims to explore the similarity between deep neural networks and human perception by comparing twelve different networks, including both general object classifiers and FER-specific models.
We employ an innovative global explainable AI method to generate heatmaps, revealing crucial facial regions for the twelve networks trained on six facial expressions. We assess these results both quantitatively and qualitatively, comparing them to ground truth masks based on Friesen and Ekman's description and among them. We use Intersection over Union (IoU), precision, recall, F1 score and normalized correlation coefficients for comparisons.
We generate 72 heatmaps to highlight critical regions for each expression and architecture. Qualitatively, models with inherently similar architectures tend to show more similarity in heatmaps, showing some learning resemblance. Specifically, eye and nose areas influence certain facial expressions, while the mouth is consistently important across all models and expressions. Quantitatively, we observe low average IoU (0.2476) and F1 score (0.3806) values across all expressions and architectures. The best-performing architecture achieves an average IoU of 0.3128 and an F1 score of 0.4622, while the worst-performing one averages 0.2086 for IoU and 0.3373 for F1 score. Additionally, there is a clear tendency for higher precision than recall in some models and expressions. 
Dendrograms, built with the normalized correlation coefficient, reveal two main clusters for most expressions: low depth models not using pre-trained weights and deeper models using them.
Findings suggest limited alignment between human and AI facial expression recognition, with network architectures influencing the similarity, as similar architectures prioritize similar facial regions.}




\maketitle

\section{Introduction}

Facial expressions are a rich and salient source of information for human communication. While facial expressions are not emotions themselves, they serve as a visual representation of an individual's emotional state as they offer observable cues that can convey emotions effectively.
Their universality has been a topic of debate \cite{Barrett}: some argue that certain facial expressions may vary across cultures or contexts, suggesting that their interpretation may not be entirely universal. However, a significant body of research builds upon the work of Ekman, who identified six fundamental and universally recognized facial expressions: anger, happiness, surprise, disgust, sadness, and fear \cite{ekmanuniversal}.
This set of facial expressions do not convey the wide range of different expressions humans can do, but they have been the  basis for understanding and interpreting emotional states in research. 

The affective computing market trends exhibit  a market growth rate of 28.2\% during 2024-2032, expecting the market to reach US\$ 682.2 billion by the end of the period \cite{Group2023}. Therefore, research on automatic Facial Expression Recognition (FER) is a very active field due to its practical application in areas such as medical diagnosis and treatment \citep{Grabowski2019}, human behavior analysis  \citep{Barreto2017, Shen2022}, or  human-computer interaction \citep{Medjden2020, Ramis2020}. 
The development of FER systems has been a topic of interest for computer vision experts, who have looked to human vision research for insights into the visual perception process. In the early stages of FER research, emphasis was placed on extracting facial features or analyzing facial appearance \citep{Pantic2000, Fasel2003}. However, in recent years, significant progress has been made by leveraging deep learning techniques \citep{Li2020, MELLOUK2020}.

Deep learning approaches, such as Convolutional Neural Networks (CNNs), have shown remarkable results in FER tasks and achieved state-of-the-art performance.
Their success have raised questions as to whether the models work like human vision,  with numerous researchers pointing out the similarities \cite{Kubilius, Mehrer, Chen2023}.

Regarding these systems, humans often exhibit cognitive anthropomorphism \citep{muellerAntro}, wherein they tend to project human-like qualities and expectations onto AI systems, assuming that AI possesses similar characteristics to human intelligence. Since 2020, literature has shown a growing research interest in anthropomorphism, highlighting its significant influence on the perception, adoption, and continued use of AI-enabled technology\cite{lisuh}.
However, it is important to note that while a deep learning approach may yield high accuracy in recognizing facial expressions, it does not necessarily imply that the underlying decision-making process resembles that of humans \citep{Borowski2019, muellerAntro}. 
Therefore, we find studies investigating similarities between deep learning approaches and human vision in the FER field and other visual human tasks \citep{Fu2023, app14062648, NEURIPS2023_f37aba0f, Geirhos2017, Kheradpisheh2016}, but thorough comparisons of human and AI behavior are still relatively rare. In fact, Bowers et al. \cite{Bowers}  state that to develop biologically plausible models of human vision, attention needs to be directed to explain psychological findings. Further, research also advocates for the exchange of knowledge between neuroscience and AI too find the most suitable computational models that resemble the information processing of the human brain \cite{hassabis}.

Mueller \citep{muellerAntro} examined known properties of human classification (e.g. representation, processing), in comparison with image classifier systems and described design strategies to cope with the differences: change the AI, change the human (e.g. instructing and training humans' expectations) or change the interaction between human and AI (e.g. showing explanations). In the case of changing the AI, mimicking human behavior may help avoiding errors and inconsistencies in performance when compared with humans and improve the deep networks by incorporating critical information for recognition \citep{Jacob2021, Ullman2016}.

In the case of human perception, the Facial Action Coding System (FACS) \citep{ekman1978manual} describes anatomically all visually discernible facial movement by defining Action Units (AUs), which are the actions of individual muscles or groups of muscles. Observing and coding a selection of AUs, EMotion FACS (EMFACS), humans can identify prototypical facial expressions that have been found to suggest certain emotions. Mining the literature, we find deep learning models trained with salient regions for human vision \citep{Khan2013} or AUs \citep{Benitez-Quiroz2017_ICCV, Pham2019} which achieve similar results as human coders with frontal faces and controlled conditions, but the accuracy decreases to below 83\% when conditions are not constrained \citep{Benitez-Quiroz2017_EmotionNet}. 

In this particular study, the focus lies on deep networks trained with facial expression images, primarily because these networks exhibit the ability to handle the challenging conditions of recognition in \textit{in the wild} scenarios \citep{Li2020, xu2022adversarial, xu2023uncertainty}. Furthermore, the study specifically deals with still images, rather than dynamic sequences or videos. We seek to answer whether deep networks observe facial AUs regions as defined in EMFACS and how human-like, in terms of visual perception of physical cues, is their processing.

To gain a deeper understanding of the inner workings of deep learning techniques, the integration of eXplainable Artificial Intelligence (XAI) techniques has emerged as a valuable approach \citep{BARREDOARRIETA202082, Adadi}. XAI techniques aim to provide explanations on how deep learning models make decisions and generate outputs, enabling human users to comprehend and interpret the results effectively \citep{Gunning2019}. XAI emphasizes on developing AI models that are interpretable and transparent \cite{speith} \cite{BARREDOARRIETA202082}. The interpretability can be achieved both by developing directly interpretable models (e.g. a decision tree) while maintaining high performance levels or using post-hoc explanations applied on trained models (specially black-box models). In this context, by applying post-hoc XAI techniques, we will analyze the human-like perception of deep learning models and explore their similarities with the FACS.

The aim of this work is twofold. On the one hand, we investigate on the similarities of deep networks with facial AUs using explainability techniques to assess how human-like are the deep networks regarding visual perceivable features. On the other hand, we compare different CNNs to study whether they focus on the same regions. Comparisons among the 12 CNNs would clarify whether FER relies on specific zones or on the architecture of the network.

Section 2 describes related works that compare human perception and deep learning techniques regarding FER. Section 3 explains the development of the FER system describing the models, the training datasets or the training process. Section 4 describes the global XAI technique proposed for explanation purposes, the ground-through mask based on Friesen and Ekman's work \cite{Friesen1983} , the metrics used to compare and the procedure followed. Section 5 presents the results and they are discussed in Section 6. Finally, the last Section concludes the work.

\section{FER comparison between human perception and deep learning techniques}
\label{sec:fer_comparison}

There is a scarce number of works comparing thoroughly human perception and deep learning techniques regarding FER.
Mining the literature, we find works focused on applying XAI techniques to the FER domain \citep{Kandeel2021,Weitz2019,ijimai,ramisInteraccion2021, Sabater-Garriz, Schiller2020,Heimerl2020}, but  it is out of their scope to explore in depth the similarities or differences regarding the human perception.

Regarding works related to analyzing how human-like are the learned features of deep learning techniques when not trained specifically with them (e.g. using AUs to train the model), Khorrami et al. \citep{Khorrami2015} already considered whether the neural networks learned facial AUs in FER. They proposed an approach to interpret which portions of the face influenced the CNN’s predictions and applied it to the extended Cohn-Kanade (CK+) dataset \citep{ck}  and the Toronto Face Dataset (TFD) \cite{Susskind}. They visualized the spatial patterns that maximally excite different filters in the convolutional layers of the network (10 selected filters in the conv3 layer). Then, they compared if the facial observed AUs aligned with facial movements by using the KL divergence on the FAU labels given in the CK+ dataset. Their results showed that CNNs are able to model high-level features that correspond to facial AUs.

Prajod et al. \citep{Prajod2021} also related the automatic FER with AUs. They presented a study on the effects of transfer learning for automatic FER from emotions to pain using the VGG16 architecture and two datasets: AffectNet for emotions and UNBC-McMaster for pain \cite{unbc}. They applied Layer-wise Relevance Propagation (LRP) saliency maps to visually compare and understand the most influent regions for the classification, both for emotion recognition and for pain recognition, and map those regions with AUs. The results showed that specific AUs related to the facial expressions of contempt and surprise were not relevant for pain recognition. 

Deramgozin et al. \citep{Deramgozin2021} presented a hybrid framework composed of a main functional pipeline with a CNN to classify input images and an explainable pipeline including a LIME visualization part. It also included a facial AUs extraction part to identify the AUs and a Multi-Layer Perceptron (MLP) classifier which reused the results of the facial AU extractor to reinforce the results obtained by the main functional pipeline. The 13 AUs used in this work were: AU01, AU02, AU05, AU06, AU07, AU09, AU12, AU15, AU17, AU20, AU23 and AU26. They evaluated their CNN and the FAU+MLP system with the CK+ dataset, and they did a visual comparison of the LIME and AUS extracted. The AUs extracted were consistent with Ekman’s basic facial expressions, although they found some differences due to the dataset and the compound emotional categories.  

Gund et al. \citep{gund2020} analyzed superficially how different spatial regions and landmark points changed in position over time. They classified expressions using two convolutional networks  trained with the CK+ dataset and tested with the SAMM dataset \citep{davison2018samm}. They highlighted the regions of interest with Class Activation Maps (CAM) and visually compared the AUs of each emotion with the CAM for landmarks for some images. 

Zhou et al. \citep{zhouEmerged} did not use XAI techniques, but they found that the pre-trained VGG-Face, could spontaneously generate expression-selective units. The accuracy of the system was far lower than humans, but it exhibited hallmarks of human expression recognition (e.g., facial expression confusion and categorical perception). However, the expression-selective units did not exhibit reliable human-like characteristics of facial expression perception. 

None of the works reviewed carry out an in-depth comparison between the learning of the model and visually perceptible features observed by humans. Since there are insights both indicating the similarity and the differences between the model learning and the human perception, in this work we will carry out a thorough quantitative comparison and a qualitative exploration.

\section{Facial Expression Recognition}

\subsection{Datasets}
\label{sec:datasets}
\begin{figure}[h]%
    \centering
    \includegraphics[width=0.8\textwidth]{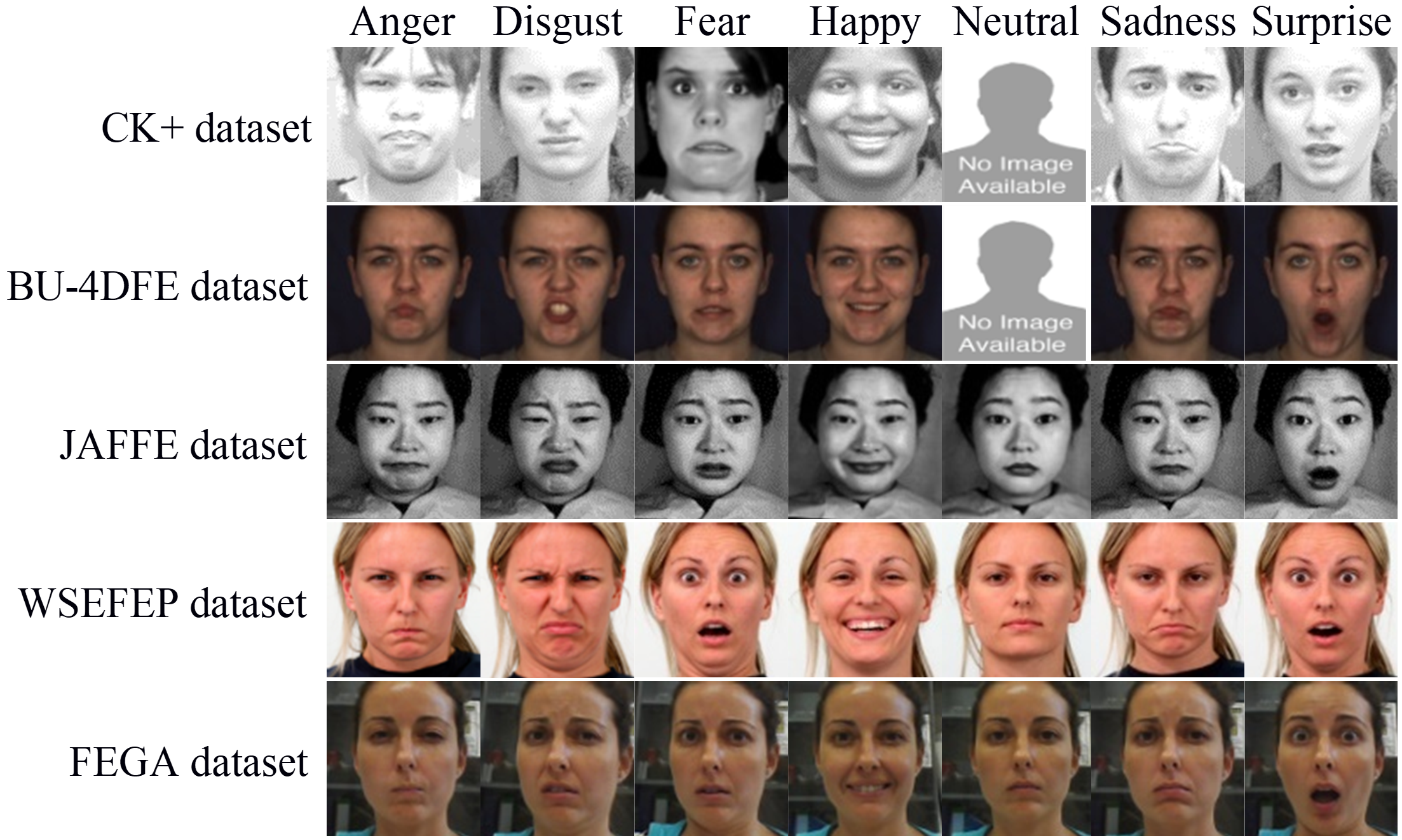}
    \caption{Samples for each class (by columns) available from each of the datasets (by rows) used in this study.}
    \label{fig:datasets}
\end{figure}

We list four standard datasets widely used in facial expression studies: the Extended Cohn-Kanade (CK+) dataset \citep{ck}, the BU-4DFE dataset \citep{BU}, the JAFFE dataset \citep{JAFFE} and the WSEFEP dataset \citep{wsefep}. In addition, we include the FEGA dataset introduced in \citep{SilNet2022}. Some samples from the different datasets can be observed in Figure \ref{fig:datasets}.

The Extended Cohn-Kanade (CK+) dataset \citep{ck} contains 593 facial expression sequences from 123 subjects ranging from 18 to 30 years old. These sequences are labelled  with one of the 7 emotions (anger, contempt, disgust, fear, happiness, sadness, and surprise) based on the expression of each subject.

The BU-4DFE dataset \citep{BU} contains 606 3D facial expression sequences from 101 subjects (58 females and 43 males). Each subject performed six sequences, one for each facial expression (anger, disgust, fear, happiness, sadness and surprise). 

The JAFFE \citep{JAFFE} dataset contains 213 images from 10 female Japanese actresses posing six basic facial expressions (happiness, surprise, fear, sadness, anger, disgust) and the neutral facial expression. 

The WSEFEP \citep{wsefep} dataset contains 210 high-quality images of 30 subjects (14 men and 16 women) posing the six basic facial expression plus the neutral face. 

Finally, the Facial Expression, Gender and Age (FEGA) dataset \citep{SilNet2022} contains 51 subjects (21 females and 30 males) ranging from 21 to 66 years old. Each subject posed six basic facial expressions plus the neutral face. Each subject was also labelled with the age and gender. 

\subsection{Pre-processing and data augmentation}
\label{sec:preprocessing}
To prepare the input of the datasets used for the training, we need to homogenize the images regarding resolution, color space and face alignment \citep{SilNet2022}. On the one hand, when training and testing, we prepare the images applying these three steps: 

\begin{enumerate}
    \item Detect the face on the image using the method proposed by Lisani et al. \citep{Lisani2017}.
    \item Align the face using the 68 facial landmarks proposed by Kazemi and Sullivan \citep{kazemi2014one} to locate the eyes position and calculate the distance between them. This method was selected for its speed, accuracy and robustness. Further, it is a very well-known technique \cite{Wu2019}. We get the angle to rotate the image by drawing a straight line between the eyes, and then, we crop the aligned face. 
    \item Convert the image to gray-scale and resize it to fit the input of the CNN.
\end{enumerate}

On the other hand, when training, we also augment the number of images applying these two steps: 

\begin{enumerate}
    \item Add variations in terms of lighting. We use the gamma correction technique \citep{MCREYNOLDS200535}.
    \item Add variations in terms of appearance, which aim to cover for small errors in the position of the eyes during the eyes’ location detection. The variations are: translation (in both axes), crop (always preserving the eyes, nose and mouth), and mirroring.
\end{enumerate}

\subsection{Model selection}
\label{sec:networks}

We built twelve models used in the literature to classify Ekman’s six basic facial expressions: nine well-known models widely used in the computer vision and neuroscience literature (AlexNet \citep{Alexnet2012}, VGG16 \citep{simonyan2015deep}, VGG19 \citep{simonyan2015deep}, ResNet50 \citep{ResNet2015}, ResNet101V2 \citep{ResNet2015}, InceptionV3 \citep{Inception2015}, Xception \citep{Xception2017}, MobileNetV3 \citep{MobileNetV3}, and EfficientNetV2 \citep{EfficientNetV2}) and three models built specifically for FER (Ramis et al. \citep{SilNet2022}, Song et al. \citep{Song2014} and Wei et al. \citep{WeiNet2015}). We name these three works as SilNet, SongNet and WeiNet respectively.
All these CNNs were trained and tested on the datasets described in subsection \ref{sec:datasets}, which were pre-processed and augmented following the steps listed in subsection \ref{sec:preprocessing}.

AlexNet, WeiNet, SongNet, and SilNet exhibit simpler architectures compared to the other models, comprising three to five convolutional layers followed by max-pooling layers, and concluding with varying numbers of dense layers. VGG16 and VGG19 are variants sharing a base architecture, distinguished by their use of compact 3x3 convolutional filters across the network and concluding with three fully connected layers, resulting in a total of 16 and 19 layers, respectively. ResNet50 and ResNet101V2 are part of the residual network family \citep{he2016deep}, which introduced residual learning through skip connections, facilitating the creation of deeper architectures. While ResNet50 comprises 50 layers, ResNet101V2 integrates optimizations such as bottleneck layers and improved skip connections, totaling 101 layers. InceptionV3 and Xception represent advancements over the original Inception \citep{szegedy2014going}, employing Inception modules with multiple parallel convolutional operations. Notably, Xception replaces standard Inception modules with depthwise separable convolutions to capture cross-channel and spatial correlations separately, thereby enhancing efficiency and performance. MobileNet \citep{howard2017mobilenets} is a family of lightweight neural network architectures designed for efficient deployment on mobile and edge devices. Characterized by depthwise separable convolutions, MobileNet architectures strike a balance between accuracy and computational efficiency, proving suitable for resource-constrained environments. MobileNetV3 incorporates features like inverted residuals, linear bottlenecks, and squeeze-and-excitation modules to further enhance efficiency and accuracy. EfficientNetV2 represents an improvement over the original EfficientNet \citep{tan2020efficientnet}, which introduced compound scaling for simultaneous optimization of model depth, width, and resolution, resulting in improved performance and efficiency. The second version introduces refinements in scaling factors, potentially enhancing overall model performance with fewer parameters.

Table \ref{table:architectures} provides a comprehensive overview of each utilized model. By incorporating these diverse architectures and comparing their performance, we aim to gain valuable insights into the impact of architectural choices on FER tasks.

\begin{table}[h]
    \caption{Networks used in this study to classify facial expressions. The number of parameters shown corresponds to the obtained model after replacing the final fully-connected layers.}\label{table:architectures}
    \begin{tabular*}{.76\textwidth}{ll|lll}
        \toprule%
        \textbf{Ref.} & \textbf{Model} & \textbf{Image Size} & \textbf{Pre-training} & \textbf{Parameters} \\
        \midrule
        \citep{Alexnet2012} & \textbf{AlexNet} & 224x224 & No & 88.7 M \\
        \citep{WeiNet2015} & \textbf{WeiNet} & 64x64 & No & 1.7 M \\
        \citep{Song2014} & \textbf{SongNet} & 224x224 & No & 172.7 K \\
        \citep{SilNet2022} & \textbf{SilNet} & 150x150 & No & 184.9 M \\
        \citep{simonyan2015deep} & \textbf{VGG16} & 224x224 & Yes & 14.7 M \\
        \citep{simonyan2015deep} & \textbf{VGG19} & 224x224 & Yes & 20 M \\
        \citep{ResNet2015} & \textbf{ResNet50} & 224x224 & Yes & 23.6 M \\
        \citep{ResNet2015} & \textbf{ResNet101V2} & 224x224 & Yes & 42.6 M \\
        \citep{Inception2015} & \textbf{InceptionV3} & 224x224 & Yes & 21.8 M \\
        \citep{Xception2017} & \textbf{Xception} & 224x224 & Yes & 20.9 M \\
        \citep{MobileNetV3} & \textbf{MobileNetV3} & 224x224 & Yes & 3 M \\
        \citep{EfficientNetV2} & \textbf{EfficientNetV2} & 224x224 & Yes & 5.9 M \\
        \botrule
    \end{tabular*}
\end{table}

The following models utilized pre-trained weights from the ImageNet dataset: VGG16, VGG19, ResNet50, ResNet101V2, InceptionV3, Xception, MobileNetV3, and EfficientNetV2. For these models, we removed the final fully connected layers and replaced them with an average 2D pooling layer and a new fully connected layer, tailored to accommodate the specific number of facial expression classes (anger, fear, disgust, sadness, surprise, and happiness) as defined by Ekman and Friesen \citep{ekman1978manual}. Conversely, the models AlexNet, WeiNet, SongNet, and SilNet were trained from scratch exclusively using facial expression images. Notably, WeiNet, SongNet, and SilNet were purpose-built explicitly for the task of facial expression recognition.

The implementation of the models was carried out in Python, leveraging the Keras library. For the models AlexNet, WeiNet, SongNet, and SilNet, we implemented each layer step by step within the Keras framework. The remaining models were readily accessible through the Keras API, complete with their pre-trained weights from the ImageNet dataset.

\subsection{Procedure}
\label{sec:training}

    We performed a k-fold cross validation with $k=5$ to train the twelve networks on the five datasets defined in subsection \ref{sec:datasets}. To do so, we first grouped the images by users for each dataset, and then split in five the users of each dataset randomly, resulting in five partitions, each containing approximately the same amount of images from each sub-dataset. Finally, following the standard k-fold cross validation procedure, we made all five possible combinations combining four of the splits for the training and leaving the remaining one for testing.

    Before starting with the cross validation, we conducted preliminary  trainings for each network to properly determine the number of epochs for each one. We determined that 10 epochs were enough to train SilNet \citep{SilNet2022}, AlexNet \citep{Alexnet2012} and WeiNet \citep{WeiNet2015}, while MobileNetV3 \citep{MobileNetV3} and EfficientNetV2 \citep{EfficientNetV2} required 10 epochs, and SongNet \citep{Song2014} required up to 25 to get proper results. The remaining networks (i.e. VGG16, VGG19, ResNet50, ResNet101V2, InceptionV3, and Xception) only needed 2 epochs to get a good performance.

    Summing up, a total of 60 trainings were carried out: 5 for each of the 12 networks, using different splits of the datasets.

\section{Measuring what do the deep networks learn and metrics for comparison}
\label{sec:explanation}

\subsection{LIME explanations}
\label{sec:lime}

    There are numerous techniques for providing visual explanations of the predictions made by deep learning (DL) models, such as perturbation-based, feature-based, and propagation-based methods, each with its own advantages and disadvantages \cite{VANDERVELDEN2022102470}. From this broad ecosystem of XAI techniques, we selected Local Interpretable Model-agnostic Explanations (LIME) \cite{Ribeiro2016}, one of the most popular model-agnostic methods \cite{ALICIOGLU2022502}. LIME can be applied to any classifier and offers locally faithful explanations for the instance being explained. When applied to images, LIME highlights the regions that most influence the prediction for a given class.
    
    Considering that it is impractical to evaluate each pixel independently for explanation purposes, clustering techniques are generally employed to create superpixels, which group similar pixels together. For this, we used the SLIC (Simple Linear Iterative Clustering) \cite{SLIC} algorithm to compute superpixels, which were then explained using LIME.

    Given a neural network architecture $R_j$ (where $j$ is one of the twelve networks described in subsection \ref{sec:networks}), a training iteration $k$, and an image $X_i$, we obtained a classification $c_{ijk}$ corresponding to a facial expression (anger, happiness, sadness, disgust, fear, or surprise). By applying LIME, we identified the image regions of most interest to the network. The outcome was an image $L_{ijk}$, with values in the range $[0, 1]$, representing minimum and maximum importance, respectively. This image was stored in grayscale to be used in further steps.

\subsection{Standardization of the explanations}
\label{sec:standardize}

    Since the dataset images were taken from different points of view and the face proportions can vary from one user to another, the exact coordinates of facial keypoints (e.g., eyes, mouth, forehead, or nose) can also vary. Therefore, we standardized all images, processing each image so that all regions of interest correspond to the same image coordinates, and applied the same transformation to the explanation grayscale images.

\begin{figure}[H]
    \captionsetup[subfigure]{justification=centering}
     \centering
     \begin{subfigure}[b]{0.3\textwidth}
         \centering
         \includegraphics[width=\textwidth]{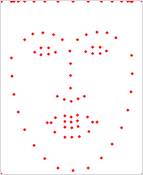}
         \caption{}
        \label{fig:landmarks}
     \end{subfigure}
     \begin{subfigure}[b]{0.3\textwidth}
         \centering
         \includegraphics[width=\textwidth]{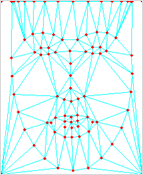}
         \caption{}
        \label{fig:triangulation}
     \end{subfigure}
    \caption{(\textbf{a}) Standard image with landmarks. (\textbf{b}) Standard image with landmarks and triangulation.}
    \label{fig:landmarks_triangulation}
\end{figure}

    The standardization process involved transforming the images using 68 landmarks estimated using the method described by Kazemi and Sullivan \citep{kazemi2014one}, along with 17 landmarks added to the top border and 4 at each corner of the image, totaling 89 landmarks. Next, we applied triangulation to each image $X_i$ using the Delaunay algorithm. The standard position of each landmark and triangle is shown in Fig. \ref{fig:landmarks_triangulation}.

\begin{figure}[H]
    \captionsetup[subfigure]{justification=centering}
     \centering
     \begin{subfigure}[b]{0.1895\textwidth}
         \centering
         \includegraphics[width=\textwidth]{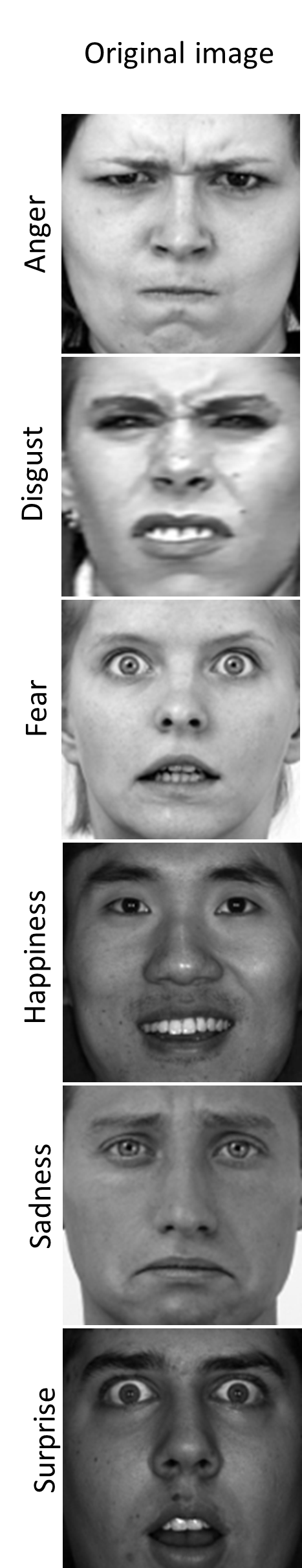}
         \caption{}
        \label{fig:norm_original}
     \end{subfigure}
     \hfill
     \begin{subfigure}[b]{0.151\textwidth}
         \centering
         \includegraphics[width=\textwidth]{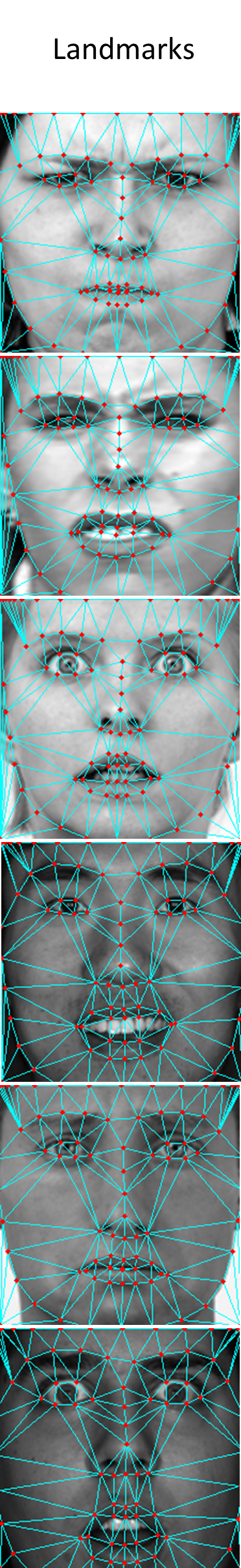}
         \caption{}
        \label{fig:norm_landmarks}
     \end{subfigure}
     \hfill
     \begin{subfigure}[b]{0.15\textwidth}
         \centering
         \includegraphics[width=\textwidth]{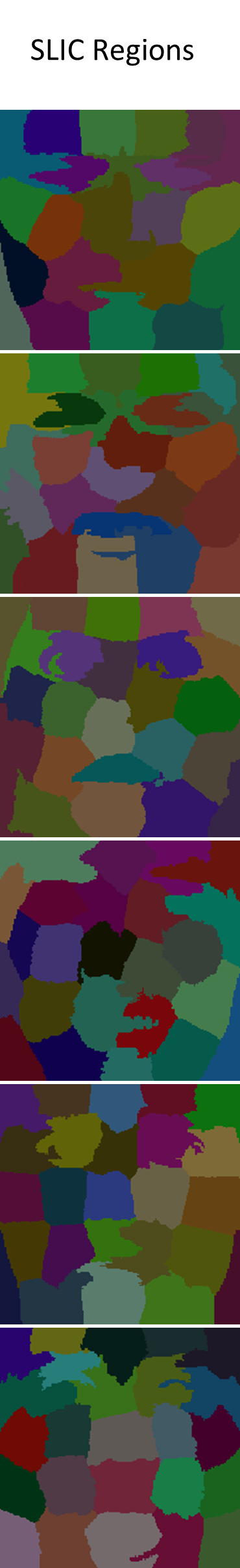}
         \caption{}
        \label{fig:norm_slic}
     \end{subfigure}
     \hfill
     \begin{subfigure}[b]{0.15\textwidth}
         \centering
         \includegraphics[width=\textwidth]{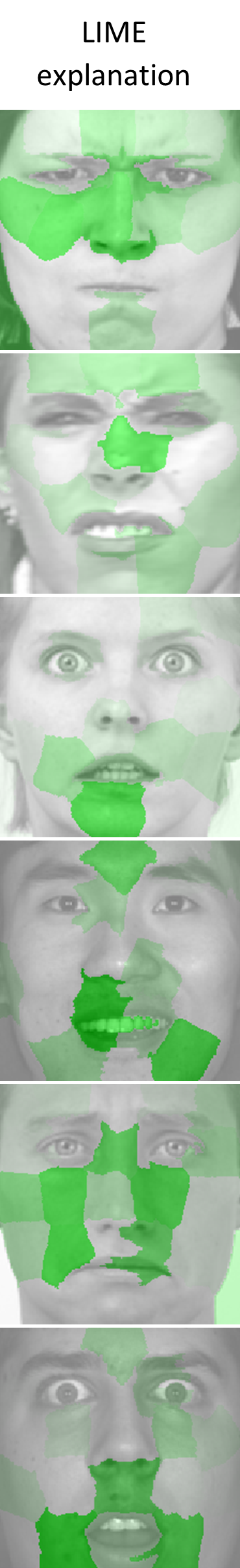}
         \caption{}
        \label{fig:norm_lime}
     \end{subfigure}
     \hfill
     \begin{subfigure}[b]{0.1303\textwidth}
         \centering
         \includegraphics[width=\textwidth]{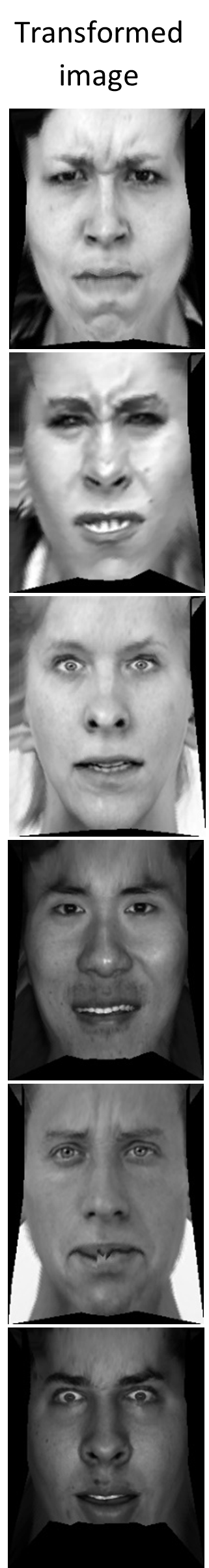}
         \caption{}
        \label{fig:norm_norm}
     \end{subfigure}
     \hfill
     \begin{subfigure}[b]{0.1253\textwidth}
         \centering
         \includegraphics[width=\textwidth]{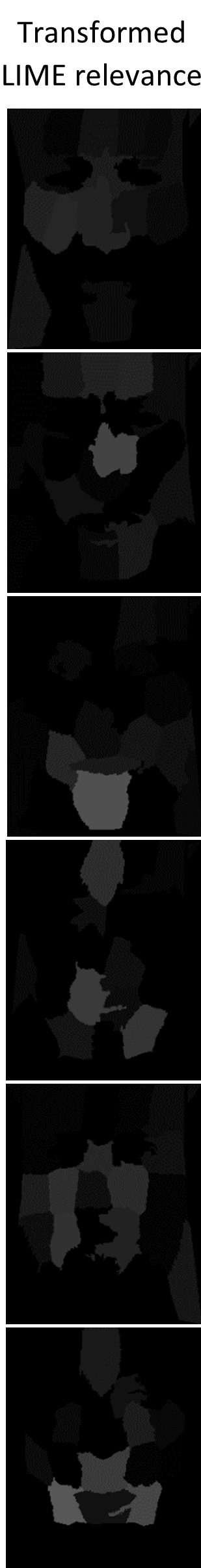}
         \caption{}
        \label{fig:norm_lime_gray}
     \end{subfigure}
    \caption{Images from the different steps involved in the explanation of an image. By rows, an example of each class: Anger, Disgust, Fear, Happiness, Sadness and Surprise. By columns: a) image being explained, b) detected face landmarks, c) superpixels computed using SLIC segmentation, d) LIME explanation, e) transformed input image using the normalized landmarks coordinates, and f) transformed LIME relevance for each region in gray scale (for further heatmap computation), using the same landmarks.}
    \label{fig:normalization}
\end{figure}

    Given an input image $X_i$ from the dataset, $L_{ijk}$ as the explanation grayscale image for the confidence of a model $j$ on a class $k$ for image $X_i$, and $S$ as the standard image, we find a transformation $T$ that moves the landmarks from $X_i$ to the same coordinates as in $S$. The transformation T is the set of affine transformations that map each triangle in the input image into another triangle in the standardized image space. Transformation T is then used to copy pixels from the input image to the standardized image. Since $X_i$ and $L_{ijk}$ share a common space (all landmarks are located at the same coordinates), the same transformation $T$ can be applied to standardize $L_{ijk}$, which we denote by $S_{ijk}$. Thus, we find $T$ knowing $S$, $X_i$, and $S = T(X_i)$, and then we apply $S_{ijk} = T(L_{ijk})$. The entire explanation and standardization process is depicted in Fig. \ref{fig:normalization}, showing the different steps for an example image from each expression.

\subsection{Importance heatmaps}
\label{sec:heatmaps}

    Once the LIME explanations were standardized, further insights could be extracted. Given a coordinate in any LIME transformed image $S_x$, the coordinate will correspond to the same face region in all images. Hence, given a network $j$, a training $k$, and a set of images $A$, we built a heatmap by calculating the average of the LIME transformed images as in Eq. \ref{eq:h}.
    
    \begin{equation}
    H=\frac{\displaystyle \sum_{X_i\in A}S_{ijk}}{card(A)}=\frac{\displaystyle \sum_{X_i\in A}{T(L_{ijk}, X_i)}}{card(A)},
    \label{eq:h}
    \end{equation}

    where $card(A)$ is the cardinality of the set $A$. By following this method, we calculated heatmaps for each network, training and facial expression, so $A_{jkc}$ represents the LIME images built with the network $j$, training set $k$ and whose classification is $c$, $c \in \{Anger,Disgust,Fear,Happiness,Sadness,Surprise\}$.

    Since we conducted five trainings for each of the twelve networks, different groupings can be done to build the heatmaps, as shown in Eq. \ref{eq:h_jc}. These groupings correspond to a summary of important regions according to the network, training set, and class.
    
    \begin{equation}
    H_{jkc} = \frac{\displaystyle \sum_{I \in A_{jkc}}{I}}{card(A_{jkc})},
    H_{jc} = \frac{\displaystyle \sum_{I \in A_{jc}}{I}}{card(A_{jc})},
    H_{c} = \frac{\displaystyle \sum_{I \in A_{c}}{I}}{card(A_{c})}, 
    \label{eq:h_jc}
    \end{equation}

    where the sum of two images is done pixel-wise. All heatmaps are images with pixel values in the range $[0,1]$, indicating the pixels' probability of being used by the network to classify the image in the facial expression $c$. Values close to 1 have more influence over the classification of the expression, while values closer to 0 are less important for the classification. $H_{jkc}$ are heatmaps summarizing the explanations of a single training (approximately 100 images per class); $H_{jc}$ group different trainings for the same network, showing the usually important regions for a model ($100 \times 5 = 500$ images per class); finally, $H_{c}$ also group the results obtained by different networks, showing the most relevant regions globally among all networks ($100 \times 5 \times 12 = 6000$ images per class).

\subsection{Ekman masks construction}
\label{sec:masks}
We build Ekman masks following the specifications defined by Friesen and Ekman \citep{Friesen1983}, to compare the influential face regions considered by the networks with anatomically visually discernible features related to emotions (i.e. EMFACS). EMFACS defines facial AUs related to facial expressions (see Table \ref{table:aus}), and each AU relates to the movement of two or more landmarks. Based on this, we build an AU-landmark related Ekman mask for each facial expression.  

\begin{table}[h]
\caption{Facial expressions described by AUs \citep{Perveen2020}.}
\label{table:aus}
\begin{tabular}{ll}
    \toprule
\textbf{Facial expression} & \textbf{Facial AUs}     \\
    \midrule
Anger                      & 4, 5, 7, 23             \\
Disgust                    & 9, 15                   \\
Fear                       & 1, 2, 4, 5, 7,   20, 26 \\
Happinness                      & 6, 12                   \\
Sadness                        & 1, 4, 15                \\
Surprise                   & 1, 2, 5, 26             \\
    \botrule
\end{tabular}
\end{table}

\begin{figure}[h]%
\centering
\includegraphics[width=0.7\textwidth]{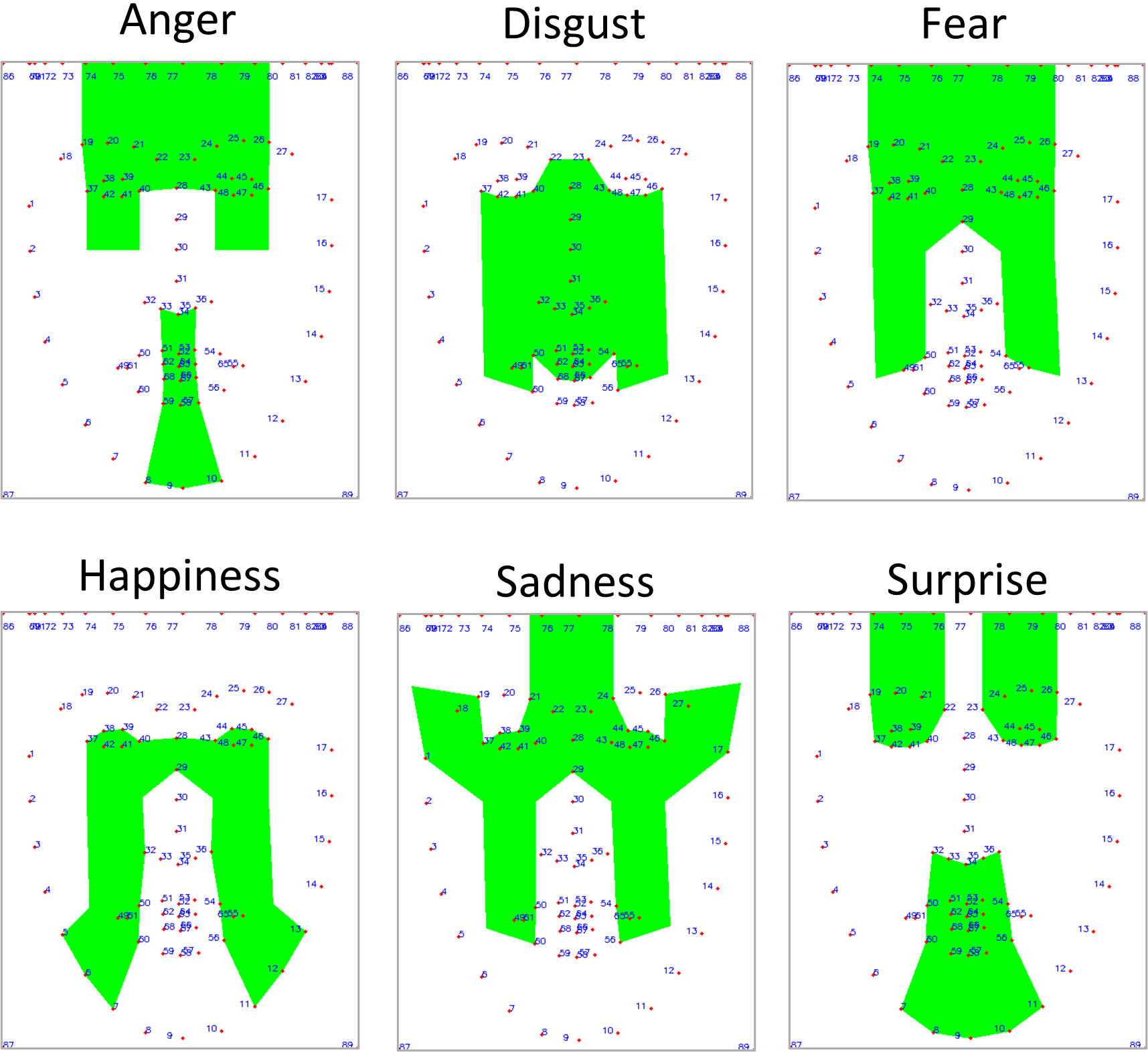}
\caption{Face areas involved in each expression, following Friesen and Ekman's \citep{Friesen1983} description.}
\label{fig:expressions}
\end{figure}

To determine the landmarks involved, we use the relations described in Perveen and Mohan’s work \citep{Perveen2020}. Based on the landmarks, we mark visually the face areas involved in the expression, to obtain an Ekman mask for every expression following Friesen and Ekman's \citep{Friesen1983} description (see Fig. \ref{fig:expressions}).

Given a facial expression, to determine the important region, the AUs are taken into account. For a given AU, the corresponding landmark is identified, or alternatively, the triangle of landmarks that includes it. If it matches a landmark, that landmark and its vicinity are marked as an important region. If the AU falls within a triangle (or on an edge), the triangle(s) that include it are marked. Subsequently, the relationship between AUs is evaluated to merge related regions.

\subsection{Metrics for heatmaps and masks comparison}
\label{sec:similarity}

To assess the obtained results and explore the computed heatmaps for the different networks, we use five different metrics: Intersection over Union (IoU), F1 score, precision, recall, and normalized correlation coefficient.

On the one hand, we use the IoU, F1 score, precision, and recall to assess the difference between two masks, where only binary data is present. This is the case when comparing thresholded heatmaps with the Ekman masks described in subsection \ref{sec:masks}. Since precision, recall, and F1 score are metrics designed to evaluate the performance of a predictor against ground truth, we consider the Ekman masks as the ground truth and the thresholded heatmaps as the predicted values. Given two binary images $I_{gt}$ (ground truth) and $I_p$ (predicted values), with equal dimensions $W \times H$ containing only values in $\{0, 1\}$ (representing the two classes: not important and important, respectively), metrics are computed as in Eq. \ref{eq:iou}, \ref{eq:precision}, \ref{eq:recall}, and \ref{eq:f1}.

\begin{equation}
IoU = \frac{ {\displaystyle \sum_{x=1}^{W} \sum_{y=1}^{H} \min(I_{gt}(x, y), I_p(x, y))} }{\displaystyle \sum_{x=1}^{W} \sum_{y=1}^{H} \max(I_{gt}(x, y), I_p(x, y))}.
\label{eq:iou}
\end{equation}

\begin{equation}
Precision = \frac{ {\displaystyle \sum_{x=1}^{W} \sum_{y=1}^{H} \min(I_{gt}(x, y), I_p(x, y))} }{\displaystyle \sum_{x=1}^{W} \sum_{y=1}^{H} I_p(x, y)}.
\label{eq:precision}
\end{equation}

\begin{equation}
Recall = \frac{ {\displaystyle \sum_{x=1}^{W} \sum_{y=1}^{H} \min(I_{gt}(x, y), I_p(x, y))} }{\displaystyle \sum_{x=1}^{W} \sum_{y=1}^{H} I_{gt}(x, y)}.
\label{eq:recall}
\end{equation}

\begin{equation}
\text{\textit{F1 score}} = \frac{2 \cdot \text{\textit{Precision}} \cdot \text{\textit{Recall}}}{\text{\textit{Precision}} + \text{\textit{Recall}}}.
\label{eq:f1}
\end{equation}

Referring to pixels with ones as white and regions with zeros as black for simplicity, precision assesses the accuracy of predicted white regions; recall evaluates how many ground truth white regions were correctly predicted; and the F1 score, the harmonic mean of precision and recall, considers both measures simultaneously, which is crucial for unbalanced problems. IoU, specifically designed for mask comparison, also provides a fair estimation in unbalanced problems.

On the other hand, to evaluate how similar are the face regions influencing the classification of facial expressions among network architectures, we use the Normalized Correlation Coefficient, which better addresses the comparison of gray-scale images. Given two heatmaps, $H_1(x,y)$ and $H_2(x,y)$, we calculate the coefficient with Eq. \ref{eq:corr}.

\begin{equation}
corr = \frac{\left\langle H_1(x,y), H_2(x,y)\right\rangle}{\sqrt{
{\left\langle H_1(x,y), H_1(x,y)\right\rangle}
{\left\langle H_2(x,y), H_2(x,y)\right\rangle}
}},
\label{eq:corr}
\end{equation}

where $\left\langle a,b\right\rangle$ denotes the dot product. This metric is 1 when images are identical, and therefore the maximum similarity, and 0 when a heatmap is the inverse of another one. This value is converted to a distance value as $1-corr$.

\subsection{Procedure}

Bearing in mind that we have five trainings for twelve networks, we start by taking a sample of 100 positives per class at random, for a total of 600 images, which will be the images to explain. We use the positives (i.e. the class predicted by the model) and not directly images taken from the class being explained because in the case that the model's prediction and the truth label do not match, we would not be explaining the model's decision. 

To follow up, we segment each image into approximately 30 regions using SLIC \citep{SLIC} and find the importance of each region using LIME, setting the number of samples to 1,000 and the background (or occlusion color) to black. This results in an explanation in the form of a gray-scale image, where brighter colors represent more relevant regions for the class being explained (see subsection \ref{sec:lime}).

The next step is to standardize the gray-scale explanation images obtaining explanations in a normalized space, where the coordinates of the landmarks have a fixed location (see subsection \ref{sec:standardize}). With all standardized explanations, we obtain heatmaps at different levels by grouping  explanations of the same class, network and training; of the same class and network (joining trainings); or only of the same class (joining trainings and networks), as seen in subsection \ref{sec:heatmaps}.

Finally, we make use of the computed heatmaps in two ways. The first one is to assess the similarity between the heatmaps and the Ekman masks described in subsection \ref{sec:masks}. To do so, it is necessary to first binarize the heatmaps, which we do by using a threshold selected using Otsu's method \citep{otsu}, and then we calculate the intersection over union. The second one is to analyze the difference among heatmaps of different networks for the same facial expression. For this, we use the normalized correlation coefficient (converted to distance) and construct different dendrograms, to visually show which networks bestow importance upon regions in a more similar way.

\section{Results}

In this section, we present a comprehensive analysis of our experimental findings. Firstly, we provide the cross-validation outcomes for the distinct neural networks employed in the study. Subsequently, we conduct an in-depth qualitative examination of the heatmaps generated by these networks. Additionally, we quantitatively assess the Intersection over Union (IoU) between the heatmaps and the Ekman masks, shedding light on the degree of similarity between the facial expression recognition capabilities of AI models and humans. Finally, we leverage dendrogram construction as an essential tool for exploring the likeness among heatmaps produced by different networks when presented with the same facial expression.

\subsection{Cross Validation}

\begin{figure}[h]%
    \centering
    \includegraphics[width=0.6\textwidth]{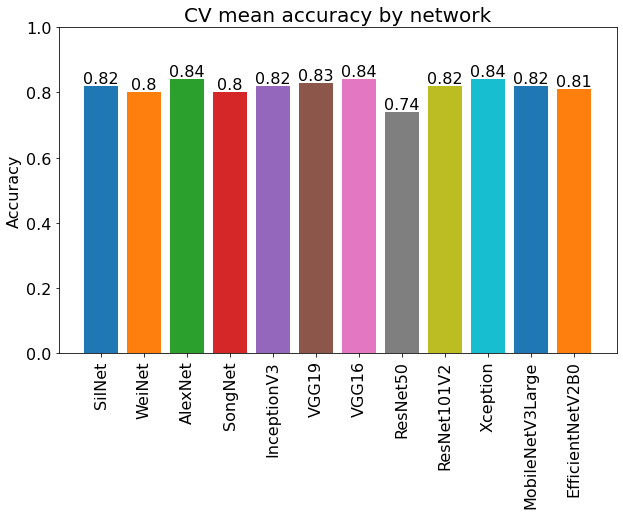}
    \caption{Mean accuracy of the cross validation for each trained network.}
    \label{fig:accuracy}
\end{figure}

To test the performance of each net, we have calculated the average accuracy across cross-validation test sets. The results are shown in Figure \ref{fig:accuracy}. As seen, the obtained results for all networks fall within the range of 80\% to 84\% accuracy, except for ResNet50, which achieved an accuracy of 74\%.

Since the aim of this study is not to maximize these results, we deem the achieved accuracy levels to be satisfactory for proceeding with the subsequent stages of the experiment: an exploration of significant facial regions identified by each network and a comparative analysis with the regions that humans perceive as important.

\subsection{Heatmaps}
\begin{figure}[H]%
    \centering
    \includegraphics[width=0.7\textwidth]{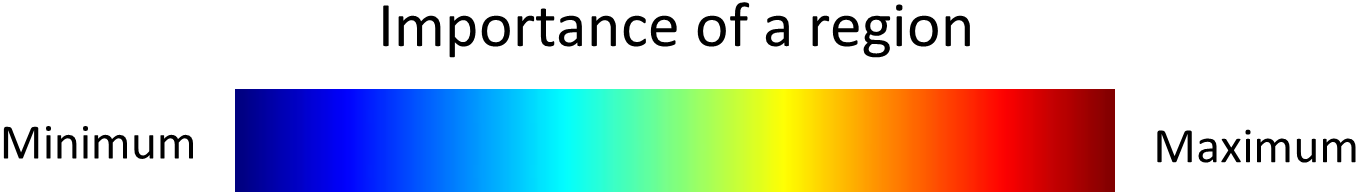}
    \caption{Color map used to represent the importance of each region in the different heatmaps.}
    \label{fig:colormap}
\end{figure}

\begin{figure}[H]%
    \centering
    \includegraphics[width=0.9\textwidth]{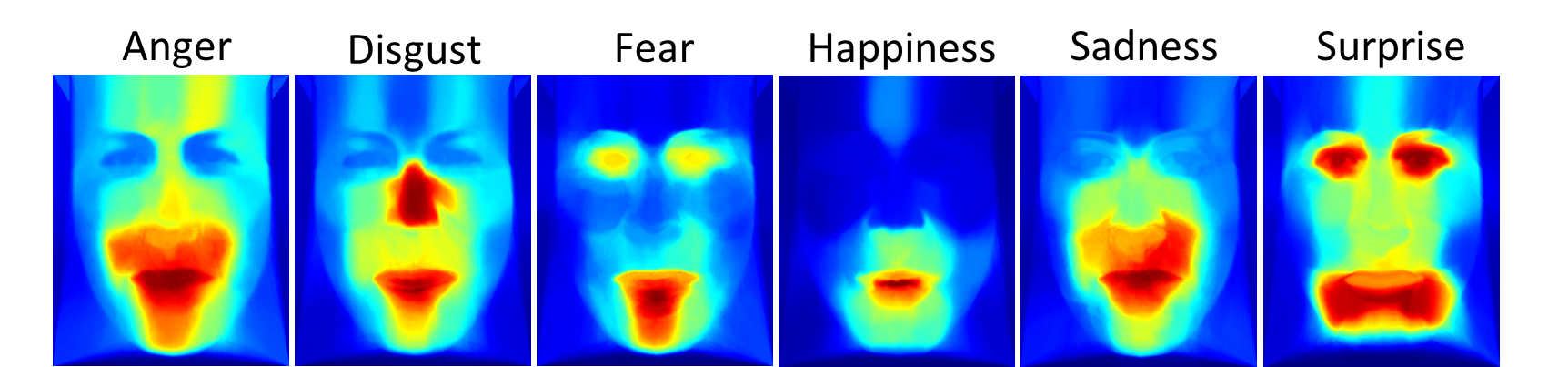}
    \caption{Resulting per-class heatmaps of the average explanations between all trained networks.}
    \label{fig:heatmaps_expressions}
\end{figure}

\begin{figure}[H]%
    \centering
    \includegraphics[width=0.8\textwidth]{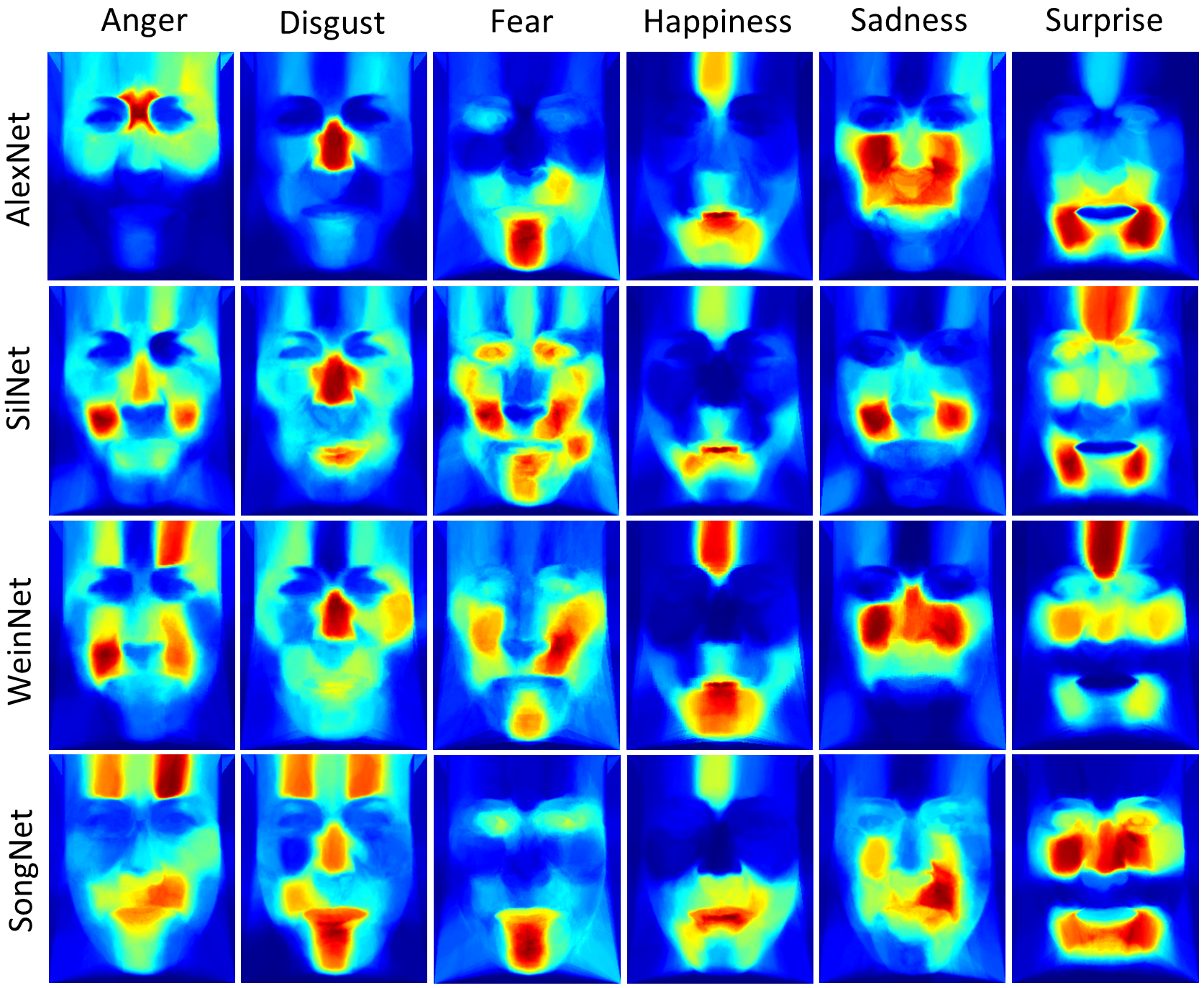}
    \caption{Resulting heatmaps of the four low depth networks not using pre-trained weights. By columns, heatmaps of the same network for the different expressions. By rows, heatmaps of the different networks for the same expression.}
    \label{fig:heatmaps1}
\end{figure}

\begin{figure}[H]%
    \centering
    \includegraphics[width=0.8\textwidth]{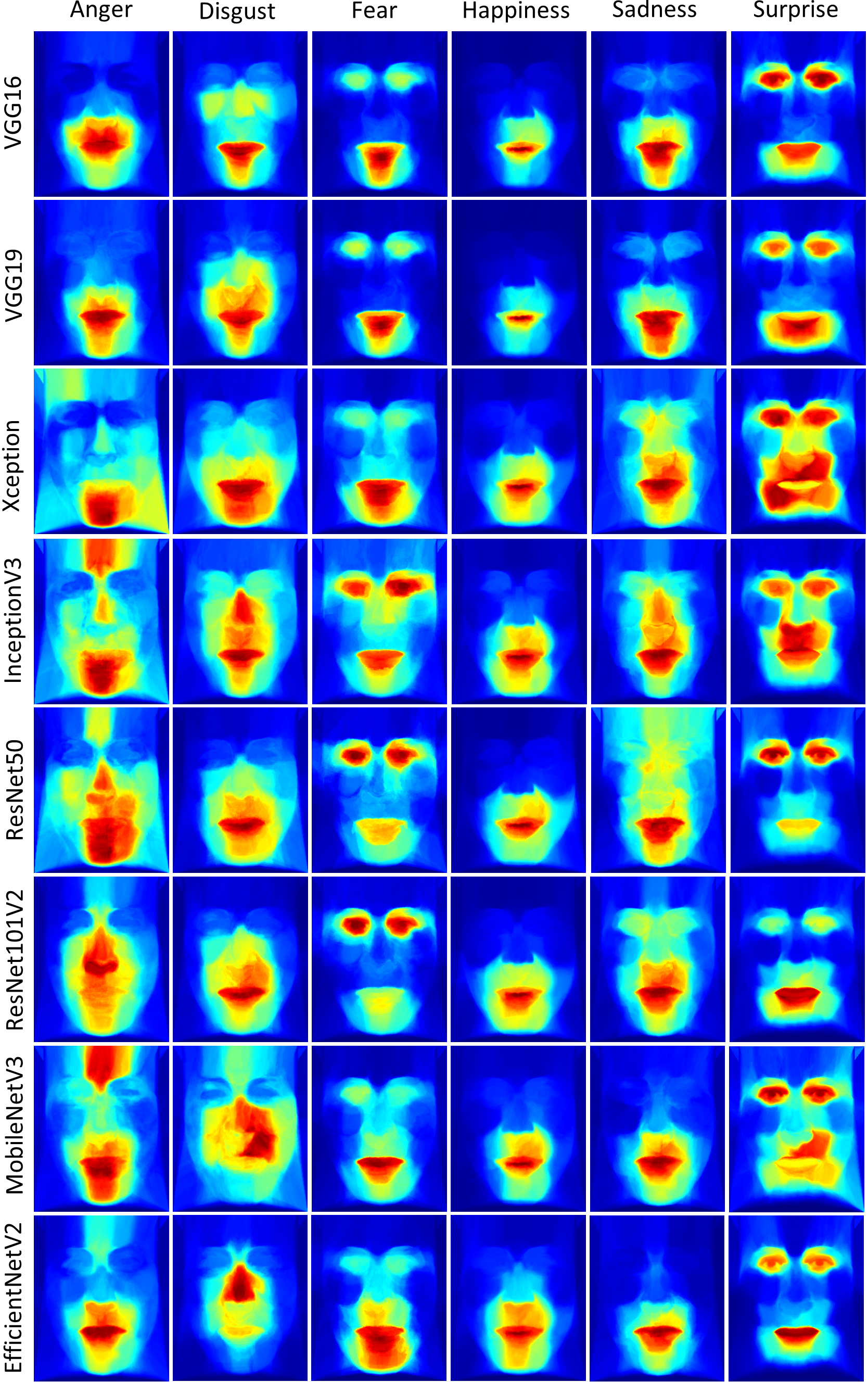}
    \caption{Resulting heatmaps of the eight deeper networks using pre-trained weights. By columns, heatmaps of the same network for the different expressions. By rows, heatmaps of the different networks for the same expression.}
    \label{fig:heatmaps2}
\end{figure}

To reflect the importance of the different regions of the face, we have employed the color map shown in Figure \ref{fig:colormap}. The resulting heatmaps of the whole explanation, standardization and summary process described in Section \ref{sec:explanation} are shown in Figures \ref{fig:heatmaps_expressions}, \ref{fig:heatmaps1} and \ref{fig:heatmaps2}.

Figure \ref{fig:heatmaps_expressions} shows the heatmaps summarizing the important regions among all networks for each expression. From these images, we can appreciate the following aspects for the different expressions:

\begin{itemize}
    \item \textbf{Anger:} the relevance is scattered through all the face, with the mouth and surroundings being the most important part. The forehead, nose, cheeks and chin can also be relevant for the models.
    
    \item \textbf{Disgust:} the nose and mouth are specially relevant for the classification, and so can be the space in between.
    
    \item \textbf{Fear:} the mouth and the chin are specially important, and the eyes can also be.
    
    \item \textbf{Happiness:} the most important part is the mouth, although the surrounding regions can also be.
    
    \item \textbf{Sadness:} the relevance is rather scattered, covering the mouth, nose and the space in between.
    
    \item \textbf{Surprise:} the important parts are the eyes and the surrounding regions of the mouth.
    
\end{itemize}

The mentioned important regions for each facial expression make sense in general compared with humans' perception: for example, the mouth should be utterly important for the happiness expression (when the person is smiling), the nose should be important for the disgust expression (wrinkle of the nose), or both the eyes and mouth should be relevant for the surprise expression (they should be wide opened).
    
Figures \ref{fig:heatmaps1} and \ref{fig:heatmaps2} display the heatmaps for a specific network and facial expression. The first one displays only low-depth networks not using pre-trained weights (i.e. SilNet, WeiNet, AlexNet and SongNet), while the second one displays deeper networks using pre-trained weights on ImageNet (i.e. VGG16, VGG19, ResNet50, ResNet101V2, InceptionV3, Xception, MobileNetV3, and EfficientNetV2). At first glance, it seems that the heatmaps in Figure \ref{fig:heatmaps1} have more scattered hot regions than Figure \ref{fig:heatmaps2}, and that there is more similarity between heatmaps of models with pre-trained weights than with models without a pre-training, which is later explored in depth in subsection \ref{sec:similarity_heatmaps}.

A specially important region for the lower depth networks appears to be the mouth, which is highlighted in almost all 48 heatmaps shown in Figure \ref{fig:heatmaps2}. The eyes are also of special importance in the recognition of fear and surprise for all networks. The nose, on the other hand, seems important for the anger, disgust and sadness expressions, although not for all networks. In the case of VGG16 and VGG19, the heatmaps are pretty alike for all expressions.

\subsection{Difference between network heatmaps and Ekman masks}
The IoU, F1 score, precision, and recall results between the Ekman masks and the thresholded heatmaps of the different networks, for each expression, are shown in Tables \ref{table:iou_f1} and \ref{table:precision_recall}. 

\begin{sidewaystable}[htbp]
\caption{IoU and F1 score between the computed heatmaps for each network and expression and the Ekman masks. Last column shows the average by network, and last row shows the average by expression. The best result for each column (and for the last row) is highlighted in bold.}
\label{table:iou_f1}
\setlength{\tabcolsep}{3pt}\begin{tabular}{l|cc|cc|cc|cc|cc|cc|cc}
\toprule
& \multicolumn{2}{c|}{\textbf{Anger}} & \multicolumn{2}{c|}{\textbf{Disgust}} & \multicolumn{2}{c|}{\textbf{Fear}} & \multicolumn{2}{c|}{\textbf{Happiness}} & \multicolumn{2}{c|}{\textbf{Sadness}} & \multicolumn{2}{c|}{\textbf{Surprise}} & \multicolumn{2}{c}{\textbf{Avg. (model)}} \\
& \textbf{IoU} & \textbf{F1 sc.} & \textbf{IoU} & \textbf{F1 sc.} & \textbf{IoU} & \textbf{F1 sc.} & \textbf{IoU} & \textbf{F1 sc.} & \textbf{IoU} & \textbf{F1 sc.} & \textbf{IoU} & \textbf{F1 sc.} & \textbf{IoU} & \textbf{F1 sc.} \\

\midrule
\textbf{SilNet} & 0.2062 & 0.3389 & 0.3841 & 0.5527 & \textbf{0.3367} & \textbf{0.5027} & 0.1141 & 0.2045 & 0.2629 & 0.4158 & 0.2785 & 0.4341 & 0.2637 & 0.4081 \\
\textbf{WeiNet} & 0.2090 & 0.3439 & 0.3042 & 0.4607 & 0.2505 & 0.3992 & 0.0988 & 0.1798 & 0.2844 & 0.4426 & 0.1380 & 0.2384 & 0.2141 & 0.3441 \\
\textbf{AlexNet} & 0.2921 & 0.4455 & 0.2329 & 0.3748 & 0.1294 & 0.2284 & 0.1151 & 0.2056 & 0.2948 & 0.4551 & 0.1874 & 0.3146 & 0.2086 & 0.3373 \\
\textbf{SongNet} & 0.3139 & 0.4706 & 0.2353 & 0.3688 & 0.1309 & 0.2220 & 0.2338 & 0.3726 & 0.2370 & 0.3795 & 0.1516 & 0.2605 & 0.2171 & 0.3457 \\
\textbf{InceptionV3} & 0.2801 & 0.4315 & \textbf{0.5923} & \textbf{0.7430} & 0.2543 & 0.4033 & 0.2297 & 0.3727 & 0.2815 & 0.4377 & 0.2387 & 0.3850 & \textbf{0.3128} & \textbf{0.4622} \\
\textbf{VGG19} & 0.1738 & 0.2842 & 0.5580 & 0.7147 & 0.1066 & 0.1855 & 0.1595 & 0.2747 & 0.1448 & 0.2434 & \textbf{0.3041} & \textbf{0.4660} & 0.2411 & 0.3614 \\
\textbf{VGG16} & 0.1231 & 0.2183 & 0.4180 & 0.5621 & 0.1112 & 0.1963 & 0.1508 & 0.2615 & 0.1834 & 0.2928 & 0.2805 & 0.4369 & 0.2112 & 0.3280 \\
\textbf{ResNet50} & 0.2112 & 0.3446 & 0.4107 & 0.5746 & 0.2063 & 0.3375 & \textbf{0.2845} & \textbf{0.4400} & \textbf{0.3119} & \textbf{0.4526} & 0.2820 & 0.4389 & 0.2844 & 0.4314 \\
\textbf{ResNet101V2} & 0.1919 & 0.3148 & 0.5048 & 0.6653 & 0.1610 & 0.2736 & 0.2611 & 0.4135 & 0.2537 & 0.3963 & 0.2817 & 0.4382 & 0.2757 & 0.4170 \\
\textbf{Xception} & 0.1052 & 0.1862 & 0.3304 & 0.4860 & 0.1694 & 0.2857 & 0.2334 & 0.3774 & 0.2311 & 0.3689 & 0.2363 & 0.3810 & 0.2176 & 0.3475 \\
\textbf{MobileNetV3} & \textbf{0.3532} & \textbf{0.5091} & 0.4980 & 0.6615 & 0.1633 & 0.2593 & 0.1966 & 0.3267 & 0.1163 & 0.2077 & 0.1902 & 0.3136 & 0.2529 & 0.3797 \\
\textbf{EfficientNetV2} & 0.2595 & 0.4016 & 0.5396 & 0.6956 & 0.1672 & 0.2720 & 0.2617 & 0.4119 & 0.1172 & 0.2045 & 0.2864 & 0.4445 & 0.2719 & 0.4050 \\
\midrule
\textbf{Avg. (express.)} & 0.2266 & 0.3574 & \textbf{0.4174} & \textbf{0.5717} & 0.1822 & 0.2971 & 0.1949 & 0.3201 & 0.2266 & 0.3581 & 0.2379 & 0.3793 & \textbf{0.2476} & \textbf{0.3806} \\
\botrule
\end{tabular}
\end{sidewaystable}

\begin{sidewaystable}[htbp]
\caption{Precision and recall between the computed heatmaps for each network and expression and the Ekman masks. Last column shows the average by network, and last row shows the average by expression. The best result for each column (and for the last row) is highlighted in bold.}
\label{table:precision_recall}
\setlength{\tabcolsep}{3pt}\begin{tabular}{l|cc|cc|cc|cc|cc|cc|cc}
\toprule
& \multicolumn{2}{c|}{\textbf{Anger}} & \multicolumn{2}{c|}{\textbf{Disgust}} & \multicolumn{2}{c|}{\textbf{Fear}} & \multicolumn{2}{c|}{\textbf{Happiness}} & \multicolumn{2}{c|}{\textbf{Sadness}} & \multicolumn{2}{c|}{\textbf{Surprise}} & \multicolumn{2}{c}{\textbf{Avg. (model)}} \\
& \textbf{Prec.} & \textbf{Recall} & \textbf{Prec.} & \textbf{Recall} & \textbf{Prec.} & \textbf{Recall} & \textbf{Prec.} & \textbf{Recall} & \textbf{Prec.} & \textbf{Recall} & \textbf{Prec.} & \textbf{Recall} & \textbf{Prec.} & \textbf{Recall} \\

\midrule
\textbf{SilNet} & 0.2942 & 0.4082 & 0.4919 & 0.6531 & 0.4836 & \textbf{0.5285} & 0.2146 & 0.2011 & 0.5574 & 0.3455 & 0.4093 & \textbf{0.4749} & 0.4085 & 0.4352 \\
\textbf{WeiNet} & 0.3271 & 0.3720 & 0.4577 & 0.5513 & 0.4481 & 0.3623 & 0.1904 & 0.1708 & \textbf{0.6167} & 0.3490 & 0.2771 & 0.2565 & 0.3862 & 0.3436 \\
\textbf{AlexNet} & 0.4044 & 0.5029 & 0.7753 & 0.2996 & 0.3114 & 0.1887 & 0.2160 & 0.1983 & 0.5070 & 0.4190 & 0.3275 & 0.3142 & 0.4236 & 0.3205 \\
\textbf{SongNet} & 0.4500 & 0.5113 & 0.3427 & 0.4042 & 0.2377 & 0.2132 & 0.3504 & 0.4017 & 0.4787 & 0.3267 & 0.2466 & 0.2776 & 0.3510 & 0.3558 \\
\textbf{InceptionV3} & 0.4327 & 0.4378 & 0.6499 & \textbf{0.8744} & 0.4235 & 0.4109 & 0.3932 & 0.3556 & 0.4670 & 0.4134 & 0.3498 & 0.4288 & 0.4527 & \textbf{0.4868} \\
\textbf{VGG19} & 0.3023 & 0.3008 & 0.6786 & 0.7725 & 0.2777 & 0.1408 & 0.3363 & 0.2326 & 0.3788 & 0.1918 & 0.5301 & 0.4169 & 0.4173 & 0.3426 \\
\textbf{VGG16} & 0.2512 & 0.1968 & 0.5692 & 0.6006 & 0.2959 & 0.1480 & 0.3438 & 0.2123 & 0.3627 & 0.2597 & \textbf{0.5335} & 0.3790 & 0.3927 & 0.2994 \\
\textbf{ResNet50} & 0.3138 & 0.3880 & 0.5231 & 0.6399 & 0.3648 & 0.3262 & \textbf{0.4290} & \textbf{0.4647} & 0.4053 & \textbf{0.5348} & 0.5313 & 0.3919 & 0.4279 & 0.4576 \\
\textbf{ResNet101V2} & 0.2880 & 0.3686 & 0.6265 & 0.7247 & \textbf{0.4937} & 0.1972 & \textbf{0.4290} & 0.4001 & 0.4618 & 0.3627 & 0.5122 & 0.4106 & \textbf{0.4685} & 0.4107 \\
\textbf{Xception} & 0.2052 & 0.1882 & 0.4623 & 0.5641 & 0.3119 & 0.2653 & 0.4067 & 0.3529 & 0.4104 & 0.3395 & 0.3448 & 0.4344 & 0.3569 & 0.3574 \\
\textbf{MobileNetV3} & \textbf{0.5075} & \textbf{0.5144} & 0.5676 & 0.8016 & 0.2659 & 0.2581 & 0.3896 & 0.2852 & 0.2776 & 0.1690 & 0.4131 & 0.3139 & 0.4036 & 0.3903 \\
\textbf{EfficientNetV2} & 0.3864 & 0.4305 & \textbf{0.8284} & 0.6226 & 0.2882 & 0.2627 & 0.4243 & 0.4048 & 0.2889 & 0.1631 & 0.5132 & 0.4115 & 0.4549 & 0.3825 \\
\midrule
\textbf{Avg. (express.)} & 0.3469 & 0.3850 & \textbf{0.5811} & \textbf{0.6257} & 0.3502 & 0.2752 & 0.3436 & 0.3067 & 0.4344 & 0.3229 & 0.4157 & 0.3759 & \textbf{0.4120} & \textbf{0.3819} \\
\botrule
\end{tabular}
\end{sidewaystable}

The IoU and F1 score demonstrate a high Pearson correlation (99\%), with F1 score values being slightly higher than IoU values. As seen in Table \ref{table:iou_f1}, the network with the highest value for one metric also achieves the highest value for the other. Among all the trained networks, InceptionV3 exhibits the best performance across all facial expressions, with IoU and F1 scores of approximately 0.31 and 0.46, respectively. It is followed closely by ResNet50, ResNet101V2, and EfficientNetV2.

When averaging by facial expressions, "Disgust" yields the most similar heatmaps to the Ekman mask, with mean values of 0.42 for IoU and 0.57 for F1 score, which are roughly two tenths higher than the values for other expressions. In particular, InceptionV3 achieves scores of 0.59 for IoU and 0.74 for F1 for this expression, the highest values among all networks and expressions.

Despite these findings, the average values for the IoU and F1 score remain relatively low across all networks (below 0.32 for IoU and 0.47 for F1 score). This indicates a significant discrepancy between the relevant regions identified by the trained networks to classify facial expressions and those defined by the constructed Ekman masks.

Table \ref{table:precision_recall} allows to appreciate higher average precision values when compared with recall for some networks and expressions. This is the case for AlexNet (10\% higher precision), VGG16 (9\%), VGG19 (7\%), EfficientNetV2 (7\%), and ResNet101V2 (6\%) networks, and for the ``sadness'' (11\%), and``fear'' (8\%) expressions. The remaining networks and expressions show an average difference between precision and recall below 5\%, and the total average precision is a 3\% higher than the total average recall. Therefore, the regions identified as important by the models tend to be correct, but they seem to miss a great proportion of them. In other words, models seem to focus on smaller regions than humans do, according to the results.

A detailed comparison between the best and worst performing models for each facial expression can be found in Table \ref{table:comparison}.

\begin{table}[]
\caption{Comparison of the best and worst performing models for each facial expression. Performance is evaluated using IoU and F1 to compare the binarized explanation heatmaps with the Ekman masks. The table lists, for each expression, the top three best and worst models along with a brief explanation of the observed differences in their performance.}
\begin{tabular}{llll}
    \toprule
    \textbf{Expression} & \textbf{Best results} & \textbf{Worst results} & \textbf{Explanation} \\
    \midrule
    
    \multirow{3}{4.5em}{\textbf{Anger}} & \multirow{3}{7em}{1.MobileNet 2.Songnet 3.AlexNet} & \multirow{3}{7em}{1.Xception 2.VGG16 3.VGG19} & \multirow{3}{20em}{Best models take into account both regions of the forehead and the mouth, while the worst performing models ignore the forehead.} \\[3em]
    
    \multirow{3}{4.5em}{\textbf{Disgust}} & \multirow{3}{7em}{1.InceptionV3 2.VGG19 3.EfficientNetV2} & \multirow{3}{7em}{1.AlexNet 2.SongNet 3.WeiNet} & \multirow{3}{20em}{Best models focus on the center of the face. Worst models also rely on other regions. AlexNet suffers from very low recall.} \\[3em]
    
    \multirow{4}{4.5em}{\textbf{Fear}} & \multirow{4}{7em}{1.ResNet101V2 2.SilNet 3.WeiNet} & \multirow{4}{7em}{1.SongNet 2.MobileNetV3 3.VGG19} & \multirow{4}{20em}{Best models correctly focus on the eyes, although they ignore regions of the forehead and around the nose. Worst models rely too much on the mouth and nose.} \\[4em]
    
    \multirow{4}{4.5em}{\textbf{Happiness}} & \multirow{4}{7em}{1.ResNet50 2.ResNet101V2 3.EfficientNetV2} & \multirow{4}{7em}{1.WeiNet 2.SilNet 3.AlexNet} & \multirow{4}{20em}{Best models put their attention solely on the mouth region, although they disregard lines of expression around the nose. Worst models also focus on the forehead.} \\[4em]
    
    \multirow{3}{4.5em}{\textbf{Sadness}} & \multirow{3}{7em}{1.WeiNet 2.SilNet 3.AlexNet} & \multirow{3}{7.5em}{1.MobileNet 2.EfficientNetV2 3.VGG16} & \multirow{3}{20em}{Best models focus on the regions under the eyes and around the nose, while the worst ones focus mainly on the mouth.} \\[3em]
    
    \multirow{3}{4.5em}{\textbf{Surprise}} & \multirow{3}{7.5em}{1.VGG19 2.EfficientNetV2 3.ResNet50} & \multirow{3}{7em}{1.WeiNet 2.SongNet 3.AlexNet} & \multirow{3}{20em}{Best models focus on the eyes and the mouth, while worst models also confer importance to the forehead and nose.} \\[2em]
    
    \botrule
\end{tabular}
\label{table:comparison}
\end{table}

\subsection{Similarity between heatmaps}
\label{sec:similarity_heatmaps}
\begin{figure}[h]
    \captionsetup[subfigure]{justification=centering}
     \centering
     \begin{subfigure}[b]{0.32\textwidth}
         \centering
         \includegraphics[width=\textwidth]{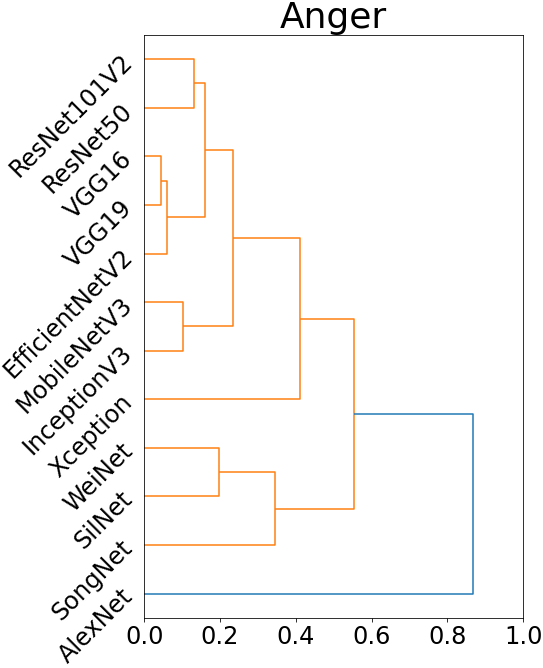}
     \end{subfigure}
     \hfill
     \begin{subfigure}[b]{0.32\textwidth}
         \centering
         \includegraphics[width=\textwidth]{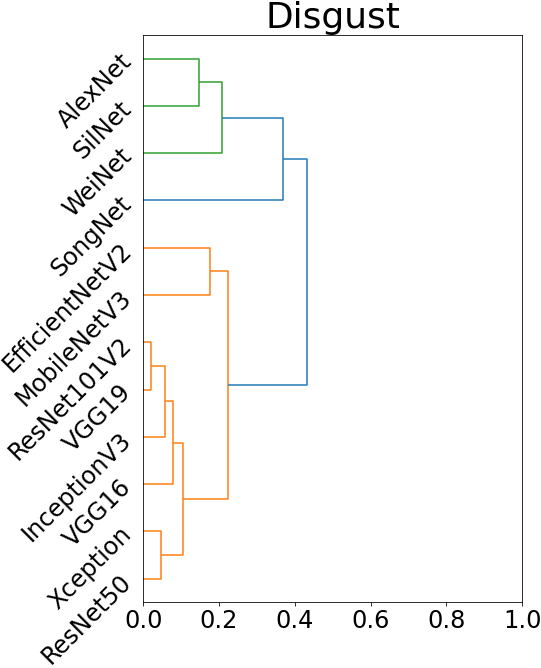}
     \end{subfigure}
     \hfill
     \begin{subfigure}[b]{0.32\textwidth}
         \centering
         \includegraphics[width=\textwidth]{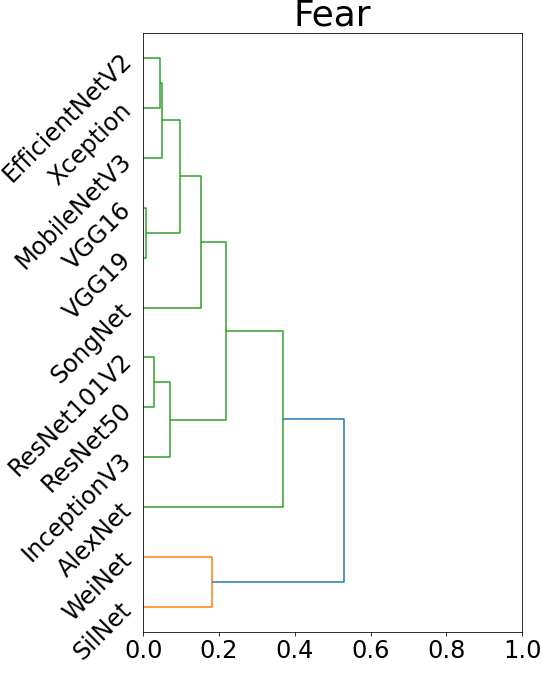}
     \end{subfigure}
     \begin{subfigure}[b]{0.32\textwidth}
         \centering
         \includegraphics[width=\textwidth]{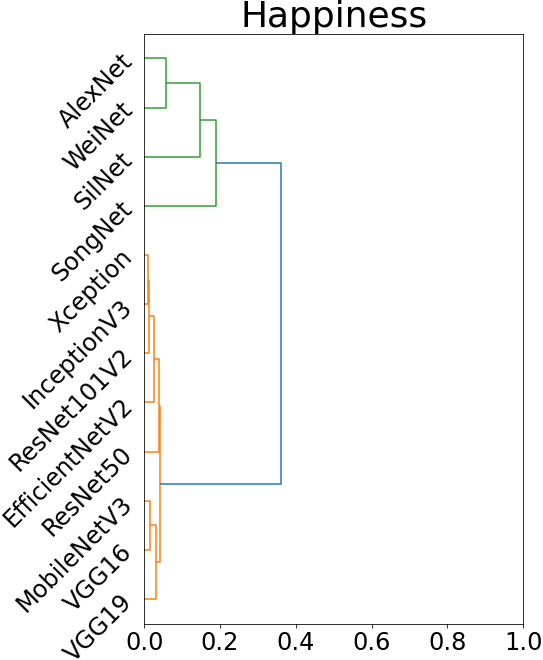}
     \end{subfigure}
     \hfill
     \begin{subfigure}[b]{0.32\textwidth}
         \centering
         \includegraphics[width=\textwidth]{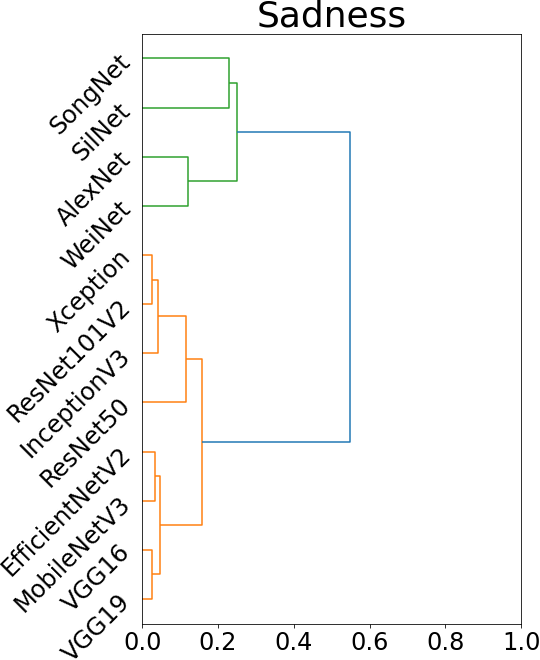}
     \end{subfigure}
     \hfill
     \begin{subfigure}[b]{0.32\textwidth}
         \centering
         \includegraphics[width=\textwidth]{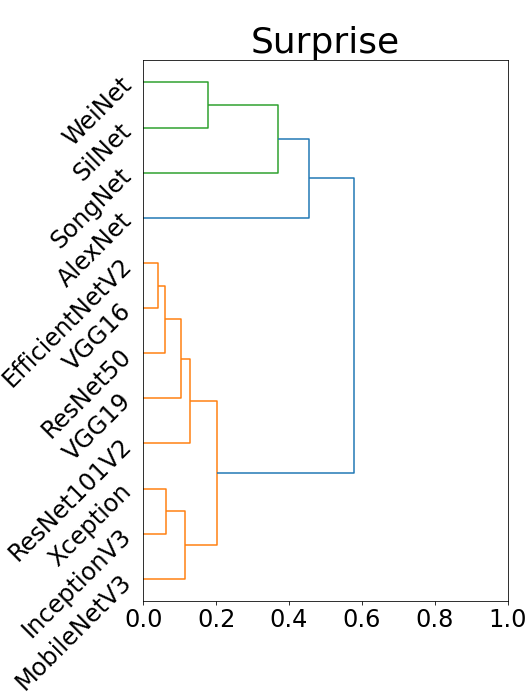}
     \end{subfigure}
    \caption{Dendrograms grouping the heatmaps of the different networks by facial expression, using the normalized correlation coefficient.}
    \label{fig:dendrograms}
\end{figure}

Figure \ref{fig:dendrograms} shows the dendrograms constructed using the normalized correlation coeficient to assess the difference between heatmaps for a network and an expression. As already seen in Figures \ref{fig:heatmaps1} and \ref{fig:heatmaps2}, there appear two main clusters for the majority of the expressions: deeper models with pre-training and low depth models with no pre-training. This seems to indicate that either the pre-training on ImageNet or the inherent similarities of deep architectures used by VGG16, VGG19, ResNet50, ResNet101V2, InceptionV3, Xception, MobileNetV3, and EfficientNetV2 lead them to converge to similar solutions for the facial expression classification problem. In addition, the difference between heatmaps for this cluster tends to be smaller than that between heatmaps of the cluster of lower depth models not using pre-trained weights (i.e. SilNet, WeiNet, AlexNet and SongNet): 0.2 vs. 0.35 in disgust, 0.05 vs. 0.2 in happiness, 0.15 vs. 0.25 in sadness, and 0.2 vs. 0.5 in surprise respectively.

A high similarity between some of the networks can also be observed. For example, SilNet and WeiNet's heatmaps are very similar for the majority of facial expressions, as is the case for VGG16 and VGG19. This indicates that similar architectures tend to generate similar heatmaps.

\section{Discussion}
    Next, we discuss the findings of the study, divided into several sections for better clarity. First, we examine the impact of the models' performance on the results. Next, we compare the similarities between different models. Then, we compare the explanation heatmaps with the Ekman masks to draw the main conclusions of the study. Finally, we discuss the implications of the results for future work and explore potential research directions.

\subsection{Performance of the models}

    Results showed consistently high accuracy across all tested models, ranging from 80\% to 84\%, with the exception of ResNet50 (74\%). We selected a broad number of networks introduced to the community between 2012 and 2022, varying in pre-training, architecture, parameters, and the purpose of design, as three of them were specifically built for facial expression recognition. As already commented, improving the performance was out of the scope of this work, but the results show that despite the diversity in design, complexity and depth, all networks demonstrated comparable accuracy levels, even ResNet50, allowing the next next steps of the study. 

    The same ensemble of datasets was used in \cite{manresa2024impact} to assess how different kinds of explanations affected users' trust in the system. The experiment involved 109 users, who achieved an accuracy of approximately 73\% on a subsample of 60 images from the dataset. Therefore, considering this accuracy to be representative, all models trained in this study outperformed humans in the facial expression classification task. This indicates the suitability of the models and the training process for this task, underscoring the importance of further investigation into their inner workings and their similarity to human reasoning. 

\subsection{Model comparison}

    When analyzing the heatmaps between deep pre-trained networks and lower depth non-pre-trained ones (see Figures \ref{fig:heatmaps1} and \ref{fig:heatmaps2}), the former focus on localized regions, while the latter consider more dispersed regions within the face. This qualitative observation is consistent with the computed dendrograms (see Figure \ref{fig:dendrograms}), where lower-depth networks cluster together in four expressions out of six (disgust, happiness, surprise, and sadness). The evident similarities within these two groups could be attributed to the presence of pre-training, the depth of the models, or both. Since the pre-training was done on the same dataset, it is possible that the models learned similar patterns that were preserved even after fine-tuning them for the new task. The lower variability observed in the group of pre-trained models compared to the non pre-trained ones, as shown in the dendograms, might indicate this phenomenon. However, this clustering could also be related to the depth of the networks, as SilNet, WeiNet, SongNet, and AlexNet are relatively shallow compared to the others. 

    Diving deeper into the similarities between networks, we observe groupings among those that share a common base architecture, such as SilNet and WeiNet, VGG16 and VGG19, and InceptionV3 and Xception. This similarity is directly evident in the computed heatmaps for most facial expressions. This correlation is likely due to their similar architecture, which causes the models to converge in a similar manner and attribute importance to the same regions of the face.

\subsection{Human-like similarities of the models}

    Regarding the important regions used by the networks for classification, we first conducted a qualitative comparison of the heatmaps. Considering all networks globally (see Figure \ref{fig:heatmaps_expressions}), we observed several patterns in how different expressions are processed. Some expressions are analyzed by the networks in a similar manner  to humans, while others focus on a subset of important regions, and still others include regions not typically considered by human perception based on Ekman's descriptions \cite{Ekman1971}. For example, the happiness expression exhibits a human-like behavior by concentrating on the lower part of the face, likely focusing on the smile. In the case of the disgust expression, the networks highlight regions important to humans, such as the nose and mouth, although the brows-forehead section is less prominent. Conversely, for expressions like anger, while the networks do pay attention to the lower part of the face, they also consider many other facial regions that are not related to AUs.

    Analyzing the quantitative results from the comparison between the models' heatmaps and the computed masks based on Ekman's descriptions (see Tables \ref{table:iou_f1} and \ref{table:precision_recall}), the findings indicate significant differences, regardless of the model architecture. This suggests that the models tend to focus on different facial regions than humans do. One major difference is the consistent emphasis on certain features across all expressions, particularly the mouth, for all deep pre-trained models. This indicates that the mouth is regarded as an important feature regardless of the expression, which does not align with Ekman's descriptions. Another notable discrepancy is the models' tendency to excessively focus  their attention on specific regions of the face, disregarding many regions considered as important for human perception.

\subsection{Implications of the findings}

Neural networks have surpassed human performance in various recognition tasks. This success has led to questions about whether these models function similarly to human vision, with researchers noting their similarities  \cite{Bowers} and proposing strategies to understand the alignment between neural networks and humans \cite{ Peterson, Muttenthaler} . The FER context is not an exception as evidenced by the existing studies described in Section \ref{sec:fer_comparison} \cite{Khorrami2015, Prajod2021, Deramgozin2021, gund2020, zhouEmerged}.

Therefore, the insights of this work have different implications. We present results showing a poor correlation between human perception and the models' rationale when classifying the facial expressions. These results are supported by different metrics and studied for diverse CNN architectures, considered among the best models of human visual object processing \cite{ Kubilius, Mehrer}. These insights can encourage the development of more human-centric neural networks to improve their performance, improve transfer, help generating human-like errors or explaining its behaviour to humans \cite{Bowers, Geirhos, Jacob2021, Ullman2016, hassabis}.

Building on these last points, humans often display cognitive anthropomorphism, that is, they tend to think AI operates in a similar way as humans. Therefore, when the FER system presents misclassifications or inexplicable decisions that seem illogical to humans, this often results in distrust towards the system, especially in high-stake domains.

Consequently, one of Mueller's strategies can be used to cope with the unexpected behavior and build trust \cite{muellerAntro}: change the AI to function in a more human-like manner, change the human (e.g. training humans' expectations) or change the interaction between human and AI (e.g. showing explanations). The strategy to use will depend on the domain and the identified requirements for the system.

Deep learning based solutions have demonstrated strong performance in FER, and although their complexity can pose challenges for user trust, it is precisely this complexity that may enable them to tackle diverse and complex challenges effectively. Consequently, efforts to enhance user trust in DL models could be achieved by changing the human or the interaction between human and AI. On the one hand, by changing the human, the system would focus on demonstrating its effectiveness and robustness, instead of making it more 'human-like'. Indeed, some studies on user trust towards DL models have shown that the accuracy of the models is more influential than the presence of explanations in building trust for tasks such as facial expression recognition \cite{manresa2024impact} . Therefore, by proving the reliability and high performance of the models could be more crucial for building user's trust than aligning them with human interpretability.

On the other hand, in critical applications like medical diagnosis, understanding how models function is crucial for their deployment in real systems. Therefore, by changing the interaction between human and AI, we would offer explanations to users on how the AI reaches specific outcomes. These explanations are essential for grasping the rationale behind the AI's decisions. Providing clear explanations helps users comprehend the AI's decision-making process, fostering confidence and enabling more informed use of these systems  \cite{guidotti2018survey}.

Finally, changing the AI aligning it with the human visual system would enable it to better meet human expectations. The outputs and decisions made by the FER system are more understandable to humans and are aligned with their mental model. This enhances transparency building trust by allowing users to understand why the system made a particular decision and it becomes easier to identify and correct errors, which is crucial for improving the reliability and accuracy of AI systems over time.

 The results of our work contribute to understand the internal working of FER recognition systems based on CNNs. This knowledge will help to further advance the field and guide the decisions on how to cope with the disparity between human perception and the system based on the problem's needs.

\section{Conclusion}

    In this study, we investigated the similarities between deep learning models and human perception in identifying the six basic facial expressions. To achieve this, twelve different networks were trained on an ensemble of five datasets for the facial expression recognition task. An explanation and standardization process was applied to uncover the important facial regions for each network and expression, represented as heatmaps. Subsequently, the obtained results were evaluated both qualitatively and quantitatively, comparing them to ground truth masks representing human perception of the expressions.

    The results reveal a significant disparity between the networks and humans in recognizing facial expressions, with values under 0.32 IoU and 0.47 F1 score in all cases. Qualitative analysis of the heatmaps, coupled with higher precision than recall for some networks and expressions, indicates a tendency of the models to overly focus on certain facial regions while neglecting others that are considered important by human perception. The consistent attribution of importance to the same regions (usually the mouth) regardless of the expression is another notable discrepancy.

    The comparison of similarity between models highlights two main clusters: one comprising pre-trained deep models and another consisting of shallower models without pre-training. Inter-cluster differences are evident in the computed heatmaps, underscoring the impact of pre-training, which may have preserved similar patterns even after fine-tuning, and the depth of the models. Additionally, architectural similarities among models contributed to smaller differences in their heatmaps, as seen with the VGG16 and VGG19 models.

    Given the negative impact that the lack of correlation between AI models and human perception has on user trust, we propose two research directions: (I) exploring how to build user trust in the models despite this discrepancy both by changing the human or the interaction, and (II) adapting the models (changing the AI) to resemble human perception while maintaining their performance.
    
    Nonetheless, we believe that research exploring and identifying similarities between CNNs and human perception, along with their systematic comparison, offers a valuable foundation for developing human-inspired computational models.
    

\section{Declarations}
\subsection{Availability of data and material}
Data of the Extended Cohn-Kanade (CK+), the BU-4DFE, the JAFFE and the WSEFEP datasets are publicly available datasets. Information access can be found in the reference.
Data of FEGA dataset is not public but is available from the authors on reasonable request.

\subsection{Competing interests}
The authors declare that they have no conflict of interest.

\subsection{Funding}
This work has been supported by Project PID2022-136779OB-C32 (PLEISAR) funded by MICIU/ AEI /10.13039/501100011033/ and FEDER, EU, Project PID2019-104829RA-I00 (EXPLAINING) funded by MCIN/ AEI /10.13039/501100011033 and Project PID2023-149079OB-I00 (EXPLAINME) funded by MICIU/AEI/10.13039/501100011033/ and FEDER, EU.
F. X. Gaya-Morey was supported by an FPU scholarship from the Ministry of European Funds, University and Culture of the Government of the Balearic Islands.

\subsection{Authors' contributions}

F.Xavier Gaya-Morey: Conceptualization, Methodology, Software, Investigation, Formal analysis, Resources, Data curation, Writing - Original Draft, Writing - Review and Editing Preparation, Visualization.

Silvia Ramis-Guarinos: Conceptualization, Methodology, Resources, Data curation, Formal analysis, Writing - Review and Editing Preparation

Cristina Manresa-Yee: Conceptualization, Methodology, Resources, Formal analysis, Writing - Original Draft,  Writing - Review and Editing Preparation, Supervision, Project administration, Funding acquisition.

Jose M. Buades-Rubio: Conceptualization, Methodology, Resources, Formal analysis, Writing - Review and Editing Preparation, Supervision, Project administration, Funding acquisition.

\subsection{Acknowledgements}
This version of the article has been accepted for publication, after peer review but is not the Version of Record and does not reflect post-acceptance improvements, or any corrections. The Version of Record is available online at: http://dx.doi.org/10.1007/s11042-024-20090-5.







\bibliography{sn-bibliography}


\begin{thebibliography}{88}
\ifx \bisbn   \undefined \def \bisbn  #1{ISBN #1}\fi
\ifx \binits  \undefined \def \binits#1{#1}\fi
\ifx \bauthor  \undefined \def \bauthor#1{#1}\fi
\ifx \batitle  \undefined \def \batitle#1{#1}\fi
\ifx \bjtitle  \undefined \def \bjtitle#1{#1}\fi
\ifx \bvolume  \undefined \def \bvolume#1{\textbf{#1}}\fi
\ifx \byear  \undefined \def \byear#1{#1}\fi
\ifx \bissue  \undefined \def \bissue#1{#1}\fi
\ifx \bfpage  \undefined \def \bfpage#1{#1}\fi
\ifx \blpage  \undefined \def \blpage #1{#1}\fi
\ifx \burl  \undefined \def \burl#1{\textsf{#1}}\fi
\ifx \doiurl  \undefined \def \doiurl#1{\url{https://doi.org/#1}}\fi
\ifx \betal  \undefined \def \betal{\textit{et al.}}\fi
\ifx \binstitute  \undefined \def \binstitute#1{#1}\fi
\ifx \binstitutionaled  \undefined \def \binstitutionaled#1{#1}\fi
\ifx \bctitle  \undefined \def \bctitle#1{#1}\fi
\ifx \beditor  \undefined \def \beditor#1{#1}\fi
\ifx \bpublisher  \undefined \def \bpublisher#1{#1}\fi
\ifx \bbtitle  \undefined \def \bbtitle#1{#1}\fi
\ifx \bedition  \undefined \def \bedition#1{#1}\fi
\ifx \bseriesno  \undefined \def \bseriesno#1{#1}\fi
\ifx \blocation  \undefined \def \blocation#1{#1}\fi
\ifx \bsertitle  \undefined \def \bsertitle#1{#1}\fi
\ifx \bsnm \undefined \def \bsnm#1{#1}\fi
\ifx \bsuffix \undefined \def \bsuffix#1{#1}\fi
\ifx \bparticle \undefined \def \bparticle#1{#1}\fi
\ifx \barticle \undefined \def \barticle#1{#1}\fi
\bibcommenthead
\ifx \bconfdate \undefined \def \bconfdate #1{#1}\fi
\ifx \botherref \undefined \def \botherref #1{#1}\fi
\ifx \url \undefined \def \url#1{\textsf{#1}}\fi
\ifx \bchapter \undefined \def \bchapter#1{#1}\fi
\ifx \bbook \undefined \def \bbook#1{#1}\fi
\ifx \bcomment \undefined \def \bcomment#1{#1}\fi
\ifx \oauthor \undefined \def \oauthor#1{#1}\fi
\ifx \citeauthoryear \undefined \def \citeauthoryear#1{#1}\fi
\ifx \endbibitem  \undefined \def \endbibitem {}\fi
\ifx \bconflocation  \undefined \def \bconflocation#1{#1}\fi
\ifx \arxivurl  \undefined \def \arxivurl#1{\textsf{#1}}\fi
\csname PreBibitemsHook\endcsname

\bibitem[\protect\citeauthoryear{Barrett et~al.}{2019}]{Barrett}
\begin{barticle}
\bauthor{\bsnm{Barrett}, \binits{L.F.}},
\bauthor{\bsnm{Adolphs}, \binits{R.}},
\bauthor{\bsnm{Marsella}, \binits{S.}},
\bauthor{\bsnm{Martinez}, \binits{A.M.}},
\bauthor{\bsnm{Pollak}, \binits{S.D.}}:
\batitle{{Emotional Expressions Reconsidered: Challenges to Inferring Emotion
  From Human Facial Movements}}.
\bjtitle{Psychological Science in the Public Interest}
\bvolume{20}(\bissue{1}),
\bfpage{1}--\blpage{68}
(\byear{2019})
\doiurl{10.1177/1529100619832930}
\end{barticle}
\endbibitem

\bibitem[\protect\citeauthoryear{Ekman}{1992}]{ekmanuniversal}
\begin{barticle}
\bauthor{\bsnm{Ekman}, \binits{P.}}:
\batitle{{An argument for basic emotions}}.
\bjtitle{Cognition and Emotion}
\bvolume{6}(\bissue{3-4}),
\bfpage{169}--\blpage{200}
(\byear{1992})
\doiurl{10.1080/02699939208411068}
\end{barticle}
\endbibitem

\bibitem[\protect\citeauthoryear{Group}{2023}]{Group2023}
\begin{botherref}
\oauthor{\bsnm{Group}, \binits{I.}}:
{Affective Computing Market Report (2024-2032). Report ID: SR112024A3711}.
Technical report,
IMARC Group
(2023).
\url{https://www.imarcgroup.com/affective-computing-market}
\end{botherref}
\endbibitem

\bibitem[\protect\citeauthoryear{Grabowski et~al.}{2019}]{Grabowski2019}
\begin{barticle}
\bauthor{\bsnm{Grabowski}, \binits{K.}},
\bauthor{\bsnm{Rynkiewicz}, \binits{A.}},
\bauthor{\bsnm{Lassalle}, \binits{A.}},
\bauthor{\bsnm{Baron-Cohen}, \binits{S.}},
\bauthor{\bsnm{Schuller}, \binits{B.}},
\bauthor{\bsnm{Cummins}, \binits{N.}},
\bauthor{\bsnm{Baird}, \binits{A.}},
\bauthor{\bsnm{Podg{\'{o}}rska-Bednarz}, \binits{J.}},
\bauthor{\bsnm{Pieni{\c{a}}{\.{z}}ek}, \binits{A.}},
\bauthor{\bsnm{{\L}ucka}, \binits{I.}}:
\batitle{{Emotional expression in psychiatric conditions: New technology for
  clinicians}}.
\bjtitle{Psychiatry and Clinical Neurosciences}
\bvolume{73}(\bissue{2}),
\bfpage{50}--\blpage{62}
(\byear{2019})
\doiurl{10.1111/pcn.12799}
\end{barticle}
\endbibitem

\bibitem[\protect\citeauthoryear{Barreto}{2017}]{Barreto2017}
\begin{barticle}
\bauthor{\bsnm{Barreto}, \binits{A.M.}}:
\batitle{{Application of facial expression studies on the field of marketing}}.
\bjtitle{Emotional expression: the brain and the face}
\bvolume{9}(\bissue{June}),
\bfpage{163}--\blpage{189}
(\byear{2017})
\end{barticle}
\endbibitem

\bibitem[\protect\citeauthoryear{Shen et~al.}{2022}]{Shen2022}
\begin{barticle}
\bauthor{\bsnm{Shen}, \binits{J.}},
\bauthor{\bsnm{Yang}, \binits{H.}},
\bauthor{\bsnm{Li}, \binits{J.}},
\bauthor{\bsnm{Cheng}, \binits{Z.}}:
\batitle{{Assessing learning engagement based on facial expression recognition
  in MOOC's scenario}}.
\bjtitle{Multimedia Systems}
\bvolume{28}(\bissue{2}),
\bfpage{469}--\blpage{478}
(\byear{2022})
\doiurl{10.1007/s00530-021-00854-x}
\end{barticle}
\endbibitem

\bibitem[\protect\citeauthoryear{Medjden et~al.}{2020}]{Medjden2020}
\begin{barticle}
\bauthor{\bsnm{Medjden}, \binits{S.}},
\bauthor{\bsnm{Ahmed}, \binits{N.}},
\bauthor{\bsnm{Lataifeh}, \binits{M.}}:
\batitle{{Adaptive user interface design and analysis using emotion recognition
  through facial expressions and body posture from an RGB-D sensor}}.
\bjtitle{PLoS ONE}
\bvolume{15}(\bissue{7}),
\bfpage{0235908}
(\byear{2020})
\doiurl{10.1371/journal.pone.0235908}
\end{barticle}
\endbibitem

\bibitem[\protect\citeauthoryear{Ramis et~al.}{2020}]{Ramis2020}
\begin{botherref}
\oauthor{\bsnm{Ramis}, \binits{S.}},
\oauthor{\bsnm{Buades}, \binits{J.M.}},
\oauthor{\bsnm{Perales}, \binits{F.J.}}:
{Using a Social Robot to Evaluate Facial Expressions in the Wild}.
Sensors
\textbf{20}(23)
(2020)
\doiurl{10.3390/s20236716}
\end{botherref}
\endbibitem

\bibitem[\protect\citeauthoryear{Pantic and Rothkrantz}{2000}]{Pantic2000}
\begin{barticle}
\bauthor{\bsnm{Pantic}, \binits{M.}},
\bauthor{\bsnm{Rothkrantz}, \binits{L.J.M.}}:
\batitle{{Automatic analysis of facial expressions: the state of the art}}.
\bjtitle{IEEE Transactions on Pattern Analysis and Machine Intelligence}
\bvolume{22}(\bissue{12}),
\bfpage{1424}--\blpage{1445}
(\byear{2000})
\doiurl{10.1109/34.895976}
\end{barticle}
\endbibitem

\bibitem[\protect\citeauthoryear{Fasel and Luettin}{2003}]{Fasel2003}
\begin{barticle}
\bauthor{\bsnm{Fasel}, \binits{B.}},
\bauthor{\bsnm{Luettin}, \binits{J.}}:
\batitle{{Automatic facial expression analysis: a survey}}.
\bjtitle{Pattern Recognition}
\bvolume{36}(\bissue{1}),
\bfpage{259}--\blpage{275}
(\byear{2003})
\doiurl{10.1016/S0031-3203(02)00052-3}
\end{barticle}
\endbibitem

\bibitem[\protect\citeauthoryear{Li and Deng}{2020}]{Li2020}
\begin{botherref}
\oauthor{\bsnm{Li}, \binits{S.}},
\oauthor{\bsnm{Deng}, \binits{W.}}:
{Deep Facial Expression Recognition: A Survey}.
IEEE Transactions on Affective Computing,
1
(2020)
\doiurl{10.1109/TAFFC.2020.2981446}
\end{botherref}
\endbibitem

\bibitem[\protect\citeauthoryear{Mellouk and Handouzi}{2020}]{MELLOUK2020}
\begin{barticle}
\bauthor{\bsnm{Mellouk}, \binits{W.}},
\bauthor{\bsnm{Handouzi}, \binits{W.}}:
\batitle{Facial emotion recognition using deep learning: review and insights}.
\bjtitle{Procedia Computer Science}
\bvolume{175},
\bfpage{689}--\blpage{694}
(\byear{2020})
\doiurl{10.1016/j.procs.2020.07.101} .
\bcomment{The 17th International Conference on Mobile Systems and Pervasive
  Computing (MobiSPC),The 15th International Conference on Future Networks and
  Communications (FNC),The 10th International Conference on Sustainable Energy
  Information Technology}
\end{barticle}
\endbibitem

\bibitem[\protect\citeauthoryear{Kubilius et~al.}{2016}]{Kubilius}
\begin{barticle}
\bauthor{\bsnm{Kubilius}, \binits{J.}},
\bauthor{\bsnm{Bracci}, \binits{S.}},
\bauthor{\bsnm{Beeck}, \binits{H.P.}}:
\batitle{{Deep Neural Networks as a Computational Model for Human Shape
  Sensitivity}}.
\bjtitle{PLOS Computational Biology}
\bvolume{12}(\bissue{4}),
\bfpage{1}--\blpage{26}
(\byear{2016})
\doiurl{10.1371/journal.pcbi.1004896}
\end{barticle}
\endbibitem

\bibitem[\protect\citeauthoryear{Mehrer et~al.}{2021}]{Mehrer}
\begin{barticle}
\bauthor{\bsnm{Mehrer}, \binits{J.}},
\bauthor{\bsnm{Spoerer}, \binits{C.J.}},
\bauthor{\bsnm{Jones}, \binits{E.C.}},
\bauthor{\bsnm{Kriegeskorte}, \binits{N.}},
\bauthor{\bsnm{Kietzmann}, \binits{T.C.}}:
\batitle{An ecologically motivated image dataset for deep learning yields
  better models of human vision}.
\bjtitle{Proceedings of the National Academy of Sciences}
\bvolume{118}(\bissue{8}),
\bfpage{2011417118}
(\byear{2021})
\doiurl{10.1073/pnas.2011417118}
{\href{https://arxiv.org/abs/https://www.pnas.org/doi/pdf/10.1073/pnas.2011417118}{{https://www.pnas.org/doi/pdf/10.1073/pnas.2011417118}}}
\end{barticle}
\endbibitem

\bibitem[\protect\citeauthoryear{Chen et~al.}{2023}]{Chen2023}
\begin{barticle}
\bauthor{\bsnm{Chen}, \binits{Y.}},
\bauthor{\bsnm{Cui}, \binits{L.}},
\bauthor{\bsnm{Ding}, \binits{M.}}:
\batitle{{Neural Processing of Affective Scenes: A Comparison between
  Convolutional Neural Networks and Human Visual Pathways}}.
\bjtitle{Journal of Vision}
\bvolume{23}(\bissue{9}),
\bfpage{5044}
(\byear{2023})
\doiurl{10.1167/jov.23.9.5044}
\end{barticle}
\endbibitem

\bibitem[\protect\citeauthoryear{Mueller}{2020}]{muellerAntro}
\begin{barticle}
\bauthor{\bsnm{Mueller}, \binits{S.T.}}:
\batitle{Cognitive anthropomorphism of ai: How humans and computers classify
  images}.
\bjtitle{Ergonomics in Design}
\bvolume{28}(\bissue{3}),
\bfpage{12}--\blpage{19}
(\byear{2020})
\doiurl{10.1177/1064804620920870}
\end{barticle}
\endbibitem

\bibitem[\protect\citeauthoryear{Li and Suh}{2021}]{lisuh}
\begin{bchapter}
\bauthor{\bsnm{Li}, \binits{M.}},
\bauthor{\bsnm{Suh}, \binits{A.}}:
\bctitle{Machinelike or humanlike? a literature review of anthropomorphism in
  ai-enabled technology}.
In: \bbtitle{Proceedings of the 54th Hawaii International Conference on System
  Sciences}.
\bsertitle{Proceedings of the Annual Hawaii International Conference on System
  Sciences},
pp. \bfpage{4053}--\blpage{4062}
(\byear{2021}).
\doiurl{10.24251/HICSS.2021.493} .
\bcomment{Research Unit(s) information for this publication is provided by the
  author(s) concerned.; 54th Hawaii International Conference on System Sciences
  (HICSS 2021), HICSS-54 ; Conference date: 04-01-2021 Through 08-01-2021}.
\burl{https://scholarspace.manoa.hawaii.edu/handle/10125/72112}
\end{bchapter}
\endbibitem

\bibitem[\protect\citeauthoryear{Borowski et~al.}{2019}]{Borowski2019}
\begin{botherref}
\oauthor{\bsnm{Borowski}, \binits{J.}},
\oauthor{\bsnm{Funke}, \binits{C.M.}},
\oauthor{\bsnm{Stosio}, \binits{K.}},
\oauthor{\bsnm{Brendel}, \binits{W.}},
\oauthor{\bsnm{Wallis}, \binits{T.S.A.}},
\oauthor{\bsnm{Bethge}, \binits{M.}}:
{The Notorious Difficulty of Comparing Human and Machine Perception},
642--646
(2019)
\doiurl{10.32470/ccn.2019.1295-0}
\end{botherref}
\endbibitem

\bibitem[\protect\citeauthoryear{Fu et~al.}{2023}]{Fu2023}
\begin{botherref}
\oauthor{\bsnm{Fu}, \binits{K.}},
\oauthor{\bsnm{Du}, \binits{C.}},
\oauthor{\bsnm{Wang}, \binits{S.}},
\oauthor{\bsnm{He}, \binits{H.}}:
Improved video emotion recognition with alignment of cnn and human brain
  representations.
IEEE Transactions on Affective Computing,
1--15
(2023)
\doiurl{10.1109/TAFFC.2023.3316173}
\end{botherref}
\endbibitem

\bibitem[\protect\citeauthoryear{Müller et~al.}{2024}]{app14062648}
\begin{botherref}
\oauthor{\bsnm{Müller}, \binits{R.}},
\oauthor{\bsnm{Dürschmidt}, \binits{M.}},
\oauthor{\bsnm{Ullrich}, \binits{J.}},
\oauthor{\bsnm{Knoll}, \binits{C.}},
\oauthor{\bsnm{Weber}, \binits{S.}},
\oauthor{\bsnm{Seitz}, \binits{S.}}:
Do humans and convolutional neural networks attend to similar areas during
  scene classification: Effects of task and image type.
Applied Sciences
\textbf{14}(6)
(2024)
\doiurl{10.3390/app14062648}
\end{botherref}
\endbibitem

\bibitem[\protect\citeauthoryear{Lee et~al.}{2023}]{NEURIPS2023_f37aba0f}
\begin{bchapter}
\bauthor{\bsnm{Lee}, \binits{J.}},
\bauthor{\bsnm{Kim}, \binits{S.}},
\bauthor{\bsnm{Won}, \binits{S.}},
\bauthor{\bsnm{Lee}, \binits{J.}},
\bauthor{\bsnm{Ghassemi}, \binits{M.}},
\bauthor{\bsnm{Thorne}, \binits{J.}},
\bauthor{\bsnm{Choi}, \binits{J.}},
\bauthor{\bsnm{Kwon}, \binits{O.-K.}},
\bauthor{\bsnm{Choi}, \binits{E.}}:
\bctitle{Visalign: Dataset for measuring the alignment between ai and humans in
  visual perception}.
In: \beditor{\bsnm{Oh}, \binits{A.}},
\beditor{\bsnm{Naumann}, \binits{T.}},
\beditor{\bsnm{Globerson}, \binits{A.}},
\beditor{\bsnm{Saenko}, \binits{K.}},
\beditor{\bsnm{Hardt}, \binits{M.}},
\beditor{\bsnm{Levine}, \binits{S.}} (eds.)
\bbtitle{Advances in Neural Information Processing Systems},
vol. \bseriesno{36},
pp. \bfpage{77119}--\blpage{77148}.
\bpublisher{Curran Associates, Inc.}, \blocation{???}
(\byear{2023}).
\burl{https://proceedings.neurips.cc/paper_files/paper/2023/file/f37aba0f53fdb59f53254fe9098b2177-Paper-Datasets_and_Benchmarks.pdf}
\end{bchapter}
\endbibitem

\bibitem[\protect\citeauthoryear{Geirhos et~al.}{2017}]{Geirhos2017}
\begin{botherref}
\oauthor{\bsnm{Geirhos}, \binits{R.}},
\oauthor{\bsnm{Janssen}, \binits{D.H.J.}},
\oauthor{\bsnm{Sch{\"{u}}tt}, \binits{H.H.}},
\oauthor{\bsnm{Rauber}, \binits{J.}},
\oauthor{\bsnm{Bethge}, \binits{M.}},
\oauthor{\bsnm{Wichmann}, \binits{F.A.}}:
{Comparing deep neural networks against humans: object recognition when the
  signal gets weaker}
(2017)
\end{botherref}
\endbibitem

\bibitem[\protect\citeauthoryear{Kheradpisheh et~al.}{2016}]{Kheradpisheh2016}
\begin{barticle}
\bauthor{\bsnm{Kheradpisheh}, \binits{S.R.}},
\bauthor{\bsnm{Ghodrati}, \binits{M.}},
\bauthor{\bsnm{Ganjtabesh}, \binits{M.}},
\bauthor{\bsnm{Masquelier}, \binits{T.}}:
\batitle{{Deep Networks Can Resemble Human Feed-forward Vision in Invariant
  Object Recognition}}.
\bjtitle{Scientific Reports}
\bvolume{6}(\bissue{1}),
\bfpage{32672}
(\byear{2016})
\doiurl{10.1038/srep32672}
\end{barticle}
\endbibitem

\bibitem[\protect\citeauthoryear{Bowers et~al.}{2023}]{Bowers}
\begin{barticle}
\bauthor{\bsnm{Bowers}, \binits{J.S.}},
\bauthor{\bsnm{Malhotra}, \binits{G.}},
\bauthor{\bsnm{Dujmović}, \binits{M.}},
\bauthor{\bsnm{Llera~Montero}, \binits{M.}},
\bauthor{\bsnm{Tsvetkov}, \binits{C.}},
\bauthor{\bsnm{Biscione}, \binits{V.}},
\bauthor{\bsnm{Puebla}, \binits{G.}},
\bauthor{\bsnm{Adolfi}, \binits{F.}},
\bauthor{\bsnm{Hummel}, \binits{J.E.}},
\bauthor{\bsnm{Heaton}, \binits{R.F.}},
\bauthor{\bsnm{al.}}:
\batitle{Deep problems with neural network models of human vision}.
\bjtitle{Behavioral and Brain Sciences}
\bvolume{46},
\bfpage{385}
(\byear{2023})
\doiurl{10.1017/S0140525X22002813}
\end{barticle}
\endbibitem

\bibitem[\protect\citeauthoryear{Hassabis et~al.}{2017}]{hassabis}
\begin{barticle}
\bauthor{\bsnm{Hassabis}, \binits{D.}},
\bauthor{\bsnm{Kumaran}, \binits{D.}},
\bauthor{\bsnm{Summerfield}, \binits{C.}},
\bauthor{\bsnm{Botvinick}, \binits{M.}}:
\batitle{Neuroscience-inspired artificial intelligence}.
\bjtitle{Neuron}
\bvolume{95}(\bissue{2}),
\bfpage{245}--\blpage{258}
(\byear{2017})
\doiurl{10.1016/j.neuron.2017.06.011}
\end{barticle}
\endbibitem

\bibitem[\protect\citeauthoryear{Jacob et~al.}{2021}]{Jacob2021}
\begin{barticle}
\bauthor{\bsnm{Jacob}, \binits{G.}},
\bauthor{\bsnm{Pramod}, \binits{R.T.}},
\bauthor{\bsnm{Katti}, \binits{H.}},
\bauthor{\bsnm{Arun}, \binits{S.P.}}:
\batitle{{Qualitative similarities and differences in visual object
  representations between brains and deep networks}}.
\bjtitle{Nature Communications}
\bvolume{12}(\bissue{1}),
\bfpage{1872}
(\byear{2021})
\doiurl{10.1038/s41467-021-22078-3}
\end{barticle}
\endbibitem

\bibitem[\protect\citeauthoryear{Ullman et~al.}{2016}]{Ullman2016}
\begin{barticle}
\bauthor{\bsnm{Ullman}, \binits{S.}},
\bauthor{\bsnm{Assif}, \binits{L.}},
\bauthor{\bsnm{Fetaya}, \binits{E.}},
\bauthor{\bsnm{Harari}, \binits{D.}}:
\batitle{{Atoms of recognition in human and computer vision}}.
\bjtitle{Proceedings of the National Academy of Sciences}
\bvolume{113}(\bissue{10}),
\bfpage{2744}--\blpage{2749}
(\byear{2016})
\doiurl{10.1073/pnas.1513198113}
\end{barticle}
\endbibitem

\bibitem[\protect\citeauthoryear{Ekman and Friesen}{1978}]{ekman1978manual}
\begin{bbook}
\bauthor{\bsnm{Ekman}, \binits{P.}},
\bauthor{\bsnm{Friesen}, \binits{W.V.}}:
\bbtitle{Manual for the Facial Action Coding System}.
\bpublisher{Consulting Psychologists Press}, \blocation{???}
(\byear{1978})
\end{bbook}
\endbibitem

\bibitem[\protect\citeauthoryear{Khan et~al.}{2013}]{Khan2013}
\begin{bchapter}
\bauthor{\bsnm{Khan}, \binits{R.A.}},
\bauthor{\bsnm{Meyer}, \binits{A.}},
\bauthor{\bsnm{Konik}, \binits{H.}},
\bauthor{\bsnm{Bouakaz}, \binits{S.}},
\bauthor{\bsnm{Khan}, \binits{R.A.}},
\bauthor{\bsnm{Meyer}, \binits{A.}},
\bauthor{\bsnm{Konik}, \binits{H.}},
\bauthor{\bsnm{Bouakaz}, \binits{S.}}:
\bctitle{{Human vision inspired framework for facial expressions recognition}}.
In: \bbtitle{Image Processing (ICIP), 2012 19th IEEE International Conference
  On, Sep 2012, Orlando, FL, United States.},
pp. \bfpage{2593}--\blpage{2596}
(\byear{2013})
\end{bchapter}
\endbibitem

\bibitem[\protect\citeauthoryear{Benitez-Quiroz
  et~al.}{2017}]{Benitez-Quiroz2017_ICCV}
\begin{bchapter}
\bauthor{\bsnm{Benitez-Quiroz}, \binits{C.F.}},
\bauthor{\bsnm{Wang}, \binits{Y.}},
\bauthor{\bsnm{Martinez}, \binits{A.M.}}:
\bctitle{{Recognition of Action Units in the Wild with Deep Nets and a New
  Global-Local Loss}}.
In: \bbtitle{2017 IEEE International Conference on Computer Vision (ICCV)},
pp. \bfpage{3990}--\blpage{3999}
(\byear{2017}).
\doiurl{10.1109/ICCV.2017.428}
\end{bchapter}
\endbibitem

\bibitem[\protect\citeauthoryear{Pham and Won}{2019}]{Pham2019}
\begin{barticle}
\bauthor{\bsnm{Pham}, \binits{T.T.D.}},
\bauthor{\bsnm{Won}, \binits{C.S.}}:
\batitle{{Facial action units for training convolutional neural networks}}.
\bjtitle{IEEE Access}
\bvolume{7},
\bfpage{77816}--\blpage{77824}
(\byear{2019})
\doiurl{10.1109/ACCESS.2019.2921241}
\end{barticle}
\endbibitem

\bibitem[\protect\citeauthoryear{Benitez-Quiroz
  et~al.}{2017}]{Benitez-Quiroz2017_EmotionNet}
\begin{botherref}
\oauthor{\bsnm{Benitez-Quiroz}, \binits{C.F.}},
\oauthor{\bsnm{Srinivasan}, \binits{R.}},
\oauthor{\bsnm{Feng}, \binits{Q.}},
\oauthor{\bsnm{Wang}, \binits{Y.}},
\oauthor{\bsnm{Martinez}, \binits{A.M.}}:
{EmotioNet Challenge: Recognition of facial expressions of emotion in the wild}
(2017)
\end{botherref}
\endbibitem

\bibitem[\protect\citeauthoryear{Xu et~al.}{2022}]{xu2022adversarial}
\begin{barticle}
\bauthor{\bsnm{Xu}, \binits{C.}},
\bauthor{\bsnm{Liu}, \binits{H.}},
\bauthor{\bsnm{Guan}, \binits{Z.}},
\bauthor{\bsnm{Wu}, \binits{X.}},
\bauthor{\bsnm{Tan}, \binits{J.}},
\bauthor{\bsnm{Ling}, \binits{B.}}:
\batitle{Adversarial incomplete multiview subspace clustering networks}.
\bjtitle{IEEE Transactions on Cybernetics}
\bvolume{52}(\bissue{10}),
\bfpage{10490}--\blpage{10503}
(\byear{2022})
\doiurl{10.1109/TCYB.2021.3062830}
\end{barticle}
\endbibitem

\bibitem[\protect\citeauthoryear{Xu et~al.}{2023}]{xu2023uncertainty}
\begin{barticle}
\bauthor{\bsnm{Xu}, \binits{C.}},
\bauthor{\bsnm{Zhao}, \binits{W.}},
\bauthor{\bsnm{Zhao}, \binits{J.}},
\bauthor{\bsnm{Guan}, \binits{Z.}},
\bauthor{\bsnm{Song}, \binits{X.}},
\bauthor{\bsnm{Li}, \binits{J.}}:
\batitle{Uncertainty-aware multiview deep learning for internet of things
  applications}.
\bjtitle{IEEE Transactions on Industrial Informatics}
\bvolume{19}(\bissue{2}),
\bfpage{1456}--\blpage{1466}
(\byear{2023})
\doiurl{10.1109/TII.2022.3206343}
\end{barticle}
\endbibitem

\bibitem[\protect\citeauthoryear{{Barredo Arrieta}
  et~al.}{2020}]{BARREDOARRIETA202082}
\begin{barticle}
\bauthor{\bsnm{{Barredo Arrieta}}, \binits{A.}},
\bauthor{\bsnm{D{\'{i}}az-Rodr{\'{i}}guez}, \binits{N.}},
\bauthor{\bsnm{{Del Ser}}, \binits{J.}},
\bauthor{\bsnm{Bennetot}, \binits{A.}},
\bauthor{\bsnm{Tabik}, \binits{S.}},
\bauthor{\bsnm{Barbado}, \binits{A.}},
\bauthor{\bsnm{Garcia}, \binits{S.}},
\bauthor{\bsnm{Gil-Lopez}, \binits{S.}},
\bauthor{\bsnm{Molina}, \binits{D.}},
\bauthor{\bsnm{Benjamins}, \binits{R.}},
\bauthor{\bsnm{Chatila}, \binits{R.}},
\bauthor{\bsnm{Herrera}, \binits{F.}}:
\batitle{{Explainable Artificial Intelligence (XAI): Concepts, taxonomies,
  opportunities and challenges toward responsible AI}}.
\bjtitle{Information Fusion}
\bvolume{58},
\bfpage{82}--\blpage{115}
(\byear{2020})
\doiurl{10.1016/j.inffus.2019.12.012}
{\href{https://arxiv.org/abs/1910.10045}{{arXiv:1910.10045}}}
\end{barticle}
\endbibitem

\bibitem[\protect\citeauthoryear{Adadi and Berrada}{2018}]{Adadi}
\begin{barticle}
\bauthor{\bsnm{Adadi}, \binits{A.}},
\bauthor{\bsnm{Berrada}, \binits{M.}}:
\batitle{Peeking inside the black-box: A survey on explainable artificial
  intelligence (xai)}.
\bjtitle{IEEE Access}
\bvolume{6},
\bfpage{52138}--\blpage{52160}
(\byear{2018})
\doiurl{10.1109/ACCESS.2018.2870052}
\end{barticle}
\endbibitem

\bibitem[\protect\citeauthoryear{Gunning and Aha}{2019}]{Gunning2019}
\begin{barticle}
\bauthor{\bsnm{Gunning}, \binits{D.}},
\bauthor{\bsnm{Aha}, \binits{D.W.}}:
\batitle{{DARPA's Explainable Artificial Intelligence {(XAI)} Program}}.
\bjtitle{{AI} Mag.}
\bvolume{40}(\bissue{2}),
\bfpage{44}--\blpage{58}
(\byear{2019})
\doiurl{10.1609/aimag.v40i2.2850}
\end{barticle}
\endbibitem

\bibitem[\protect\citeauthoryear{Speith}{2022}]{speith}
\begin{bchapter}
\bauthor{\bsnm{Speith}, \binits{T.}}:
\bctitle{A review of taxonomies of explainable artificial intelligence (xai)
  methods}.
In: \bbtitle{Proceedings of the 2022 ACM Conference on Fairness,
  Accountability, and Transparency}.
\bsertitle{FAccT '22},
pp. \bfpage{2239}--\blpage{2250}.
\bpublisher{Association for Computing Machinery},
\blocation{New York, NY, USA}
(\byear{2022}).
\doiurl{10.1145/3531146.3534639} .
\burl{https://doi.org/10.1145/3531146.3534639}
\end{bchapter}
\endbibitem

\bibitem[\protect\citeauthoryear{Friesen and Ekman}{1983}]{Friesen1983}
\begin{bchapter}
\bauthor{\bsnm{Friesen}, \binits{W.V.}},
\bauthor{\bsnm{Ekman}, \binits{P.}}:
\bctitle{{EMFACS-7: Emotional Facial Action Coding System}}.
(\byear{1983})
\end{bchapter}
\endbibitem

\bibitem[\protect\citeauthoryear{Kandeel et~al.}{2021}]{Kandeel2021}
\begin{bchapter}
\bauthor{\bsnm{Kandeel}, \binits{A.A.}},
\bauthor{\bsnm{Abbas}, \binits{H.M.}},
\bauthor{\bsnm{Hassanein}, \binits{H.S.}}:
\bctitle{{Explainable Model Selection of a Convolutional Neural Network for
  Driver's Facial Emotion Identification}}.
In: \beditor{\bsnm{{Del Bimbo}}, \binits{A.}},
\beditor{\bsnm{Cucchiara}, \binits{R.}},
\beditor{\bsnm{Sclaroff}, \binits{S.}},
\beditor{\bsnm{Farinella}, \binits{G.M.}},
\beditor{\bsnm{Mei}, \binits{T.}},
\beditor{\bsnm{Bertini}, \binits{M.}},
\beditor{\bsnm{Escalante}, \binits{H.J.}},
\beditor{\bsnm{Vezzani}, \binits{R.}} (eds.)
\bbtitle{Pattern Recognition. ICPR International Workshops and Challenges},
pp. \bfpage{699}--\blpage{713}.
\bpublisher{Springer},
\blocation{Cham}
(\byear{2021})
\end{bchapter}
\endbibitem

\bibitem[\protect\citeauthoryear{Weitz et~al.}{2019}]{Weitz2019}
\begin{barticle}
\bauthor{\bsnm{Weitz}, \binits{K.}},
\bauthor{\bsnm{Hassan}, \binits{T.}},
\bauthor{\bsnm{Schmid}, \binits{U.}},
\bauthor{\bsnm{Garbas}, \binits{J.}}:
\batitle{{Deep-learned faces of pain and emotions: Elucidating the differences
  of facial expressions with the help of explainable AI methods}}.
\bjtitle{tm - Technisches Messen}
\bvolume{86},
\bfpage{404}--\blpage{412}
(\byear{2019})
\end{barticle}
\endbibitem

\bibitem[\protect\citeauthoryear{Manresa-Yee et~al.}{}]{ijimai}
\begin{botherref}
\oauthor{\bsnm{Manresa-Yee}, \binits{C.}},
\oauthor{\bsnm{Ramis}, \binits{S.}},
\oauthor{\bsnm{Buades}, \binits{J.M.}}:
{Analysis of Gender Differences in Facial Expression Recognition Based on Deep
  Learning Using Explainable Artificial Intelligence}.
International Journal of Interactive Multimedia and Artificial Intelligence (In
  press)
\doiurl{10.9781/ijimai.2023.04.003}
\end{botherref}
\endbibitem

\bibitem[\protect\citeauthoryear{Manresa-Yee
  et~al.}{2022}]{ramisInteraccion2021}
\begin{bchapter}
\bauthor{\bsnm{Manresa-Yee}, \binits{C.}},
\bauthor{\bsnm{Ramis~Guarinos}, \binits{S.}},
\bauthor{\bsnm{Buades~Rubio}, \binits{J.M.}}:
\bctitle{Facial expression recognition: Impact of gender on fairness and
  expressions}.
In: \bbtitle{Proceedings of the XXII International Conference on Human Computer
  Interaction}.
\bsertitle{Interacci\'{o}n '22}.
\bpublisher{Association for Computing Machinery},
\blocation{New York, NY, USA}
(\byear{2022}).
\doiurl{10.1145/3549865.3549904}
\end{bchapter}
\endbibitem

\bibitem[\protect\citeauthoryear{Sabater-G{\'{a}}rriz
  et~al.}{}]{Sabater-Garriz}
\begin{botherref}
\oauthor{\bsnm{Sabater-G{\'{a}}rriz}, \binits{A.}},
\oauthor{\bsnm{Gaya-Morey}, \binits{F.X.}},
\oauthor{\bsnm{Buades}, \binits{J.M.}},
\oauthor{\bsnm{Manresa-Yee}, \binits{C.}},
\oauthor{\bsnm{Montoya}, \binits{P.}},
\oauthor{\bsnm{Riquelme}, \binits{I.}}:
{Automated facial recognition system using deep learning for pain assessment in
  adults with cerebral palsy.}
Digital Health
(In press)
\end{botherref}
\endbibitem

\bibitem[\protect\citeauthoryear{Schiller et~al.}{2020}]{Schiller2020}
\begin{barticle}
\bauthor{\bsnm{Schiller}, \binits{D.}},
\bauthor{\bsnm{Huber}, \binits{T.}},
\bauthor{\bsnm{Dietz}, \binits{M.}},
\bauthor{\bsnm{Andr{\'{e}}}, \binits{E.}}:
\batitle{{Relevance-Based Data Masking: A Model-Agnostic Transfer Learning
  Approach for Facial Expression Recognition}}.
\bjtitle{Frontiers in Computer Science}
\bvolume{2},
\bfpage{6}
(\byear{2020})
\doiurl{10.3389/fcomp.2020.00006}
\end{barticle}
\endbibitem

\bibitem[\protect\citeauthoryear{Heimerl et~al.}{2020}]{Heimerl2020}
\begin{barticle}
\bauthor{\bsnm{Heimerl}, \binits{A.}},
\bauthor{\bsnm{Weitz}, \binits{K.}},
\bauthor{\bsnm{Baur}, \binits{T.}},
\bauthor{\bsnm{Andre}, \binits{E.}}:
\batitle{{Unraveling ML Models of Emotion with NOVA: Multi-Level Explainable AI
  for Non-Experts}}.
\bjtitle{IEEE Transactions on Affective Computing}
\bvolume{1}(\bissue{1}),
\bfpage{1}--\blpage{13}
(\byear{2020})
\doiurl{10.1109/TAFFC.2020.3043603}
\end{barticle}
\endbibitem

\bibitem[\protect\citeauthoryear{Khorrami et~al.}{2015}]{Khorrami2015}
\begin{bchapter}
\bauthor{\bsnm{Khorrami}, \binits{P.}},
\bauthor{\bsnm{Paine}, \binits{T.L.}},
\bauthor{\bsnm{Huang}, \binits{T.S.}}:
\bctitle{{Do Deep Neural Networks Learn Facial Action Units When Doing
  Expression Recognition?}}
In: \bbtitle{2015 IEEE International Conference on Computer Vision Workshop
  (ICCVW)},
pp. \bfpage{19}--\blpage{27}
(\byear{2015}).
\doiurl{10.1109/ICCVW.2015.12}
\end{bchapter}
\endbibitem

\bibitem[\protect\citeauthoryear{Lucey et~al.}{2010}]{ck}
\begin{bchapter}
\bauthor{\bsnm{Lucey}, \binits{P.}},
\bauthor{\bsnm{Cohn}, \binits{J.F.}},
\bauthor{\bsnm{Kanade}, \binits{T.}},
\bauthor{\bsnm{Saragih}, \binits{J.}},
\bauthor{\bsnm{Ambadar}, \binits{Z.}},
\bauthor{\bsnm{Matthews}, \binits{I.}}:
\bctitle{The extended cohn-kanade dataset (ck+): A complete dataset for action
  unit and emotion-specified expression}.
In: \bbtitle{2010 Ieee Computer Society Conference on Computer Vision and
  Pattern Recognition-workshops},
pp. \bfpage{94}--\blpage{101}
(\byear{2010}).
\bcomment{IEEE}
\end{bchapter}
\endbibitem

\bibitem[\protect\citeauthoryear{Susskind et~al.}{}]{Susskind}
\begin{botherref}
\oauthor{\bsnm{Susskind}, \binits{J.M.}},
\oauthor{\bsnm{Anderson}, \binits{A.K.}},
\oauthor{\bsnm{Hinton}, \binits{G.E.}}:
{The Toronto Face Database.}
Technical report
\end{botherref}
\endbibitem

\bibitem[\protect\citeauthoryear{Prajod et~al.}{2022}]{Prajod2021}
\begin{bbook}
\bauthor{\bsnm{Prajod}, \binits{P.}},
\bauthor{\bsnm{Schiller}, \binits{D.}},
\bauthor{\bsnm{Huber}, \binits{T.}},
\bauthor{\bsnm{Andr{\'{e}}}, \binits{E.}}:
In: \beditor{\bsnm{Shaban-Nejad}, \binits{A.}},
\beditor{\bsnm{Michalowski}, \binits{M.}},
\beditor{\bsnm{Bianco}, \binits{S.}} (eds.)
\bbtitle{{Do Deep Neural Networks Forget Facial Action Units?---Exploring the
  Effects of Transfer Learning in Health Related Facial Expression
  Recognition}},
pp. \bfpage{217}--\blpage{233}.
\bpublisher{Springer},
\blocation{Cham}
(\byear{2022}).
\doiurl{10.1007/978-3-030-93080-6_16}
\end{bbook}
\endbibitem

\bibitem[\protect\citeauthoryear{Lucey et~al.}{2011}]{unbc}
\begin{bchapter}
\bauthor{\bsnm{Lucey}, \binits{P.}},
\bauthor{\bsnm{Cohn}, \binits{J.F.}},
\bauthor{\bsnm{Prkachin}, \binits{K.M.}},
\bauthor{\bsnm{Solomon}, \binits{P.E.}},
\bauthor{\bsnm{Matthews}, \binits{I.}}:
\bctitle{Painful data: The unbc-mcmaster shoulder pain expression archive
  database}.
In: \bbtitle{2011 IEEE International Conference on Automatic Face \& Gesture
  Recognition (FG)},
pp. \bfpage{57}--\blpage{64}
(\byear{2011}).
\doiurl{10.1109/FG.2011.5771462}
\end{bchapter}
\endbibitem

\bibitem[\protect\citeauthoryear{Deramgozin et~al.}{2021}]{Deramgozin2021}
\begin{bbook}
\bauthor{\bsnm{Deramgozin}, \binits{M.}},
\bauthor{\bsnm{Jovanovic}, \binits{S.}},
\bauthor{\bsnm{Rabah}, \binits{H.}},
\bauthor{\bsnm{Ramzan}, \binits{N.}}:
\bbtitle{A Hybrid Explainable AI Framework Applied to Global and Local Facial
  Expression Recognition},
(\byear{2021}).
\doiurl{10.1109/IST50367.2021.9651357}
\end{bbook}
\endbibitem

\bibitem[\protect\citeauthoryear{Gund et~al.}{2021}]{gund2020}
\begin{bchapter}
\bauthor{\bsnm{Gund}, \binits{M.}},
\bauthor{\bsnm{Bharadwaj}, \binits{A.R.}},
\bauthor{\bsnm{Nwogu}, \binits{I.}}:
\bctitle{Interpretable emotion classification using temporal convolutional
  models}.
In: \bbtitle{2020 25th International Conference on Pattern Recognition (ICPR)},
pp. \bfpage{6367}--\blpage{6374}
(\byear{2021}).
\doiurl{10.1109/ICPR48806.2021.9412134}
\end{bchapter}
\endbibitem

\bibitem[\protect\citeauthoryear{Davison et~al.}{2018}]{davison2018samm}
\begin{barticle}
\bauthor{\bsnm{Davison}, \binits{A.K.}},
\bauthor{\bsnm{Lansley}, \binits{C.}},
\bauthor{\bsnm{Costen}, \binits{N.}},
\bauthor{\bsnm{Tan}, \binits{K.}},
\bauthor{\bsnm{Yap}, \binits{M.H.}}:
\batitle{Samm: A spontaneous micro-facial movement dataset}.
\bjtitle{IEEE Transactions on Affective Computing}
\bvolume{9}(\bissue{1}),
\bfpage{116}--\blpage{129}
(\byear{2018})
\doiurl{10.1109/TAFFC.2016.2573832}
\end{barticle}
\endbibitem

\bibitem[\protect\citeauthoryear{Zhou et~al.}{2022}]{zhouEmerged}
\begin{barticle}
\bauthor{\bsnm{Zhou}, \binits{L.}},
\bauthor{\bsnm{Yang}, \binits{A.}},
\bauthor{\bsnm{Meng}, \binits{M.}},
\bauthor{\bsnm{Zhou}, \binits{K.}}:
\batitle{Emerged human-like facial expression representation in a deep
  convolutional neural network}.
\bjtitle{Science Advances}
\bvolume{8}(\bissue{12}),
\bfpage{4383}
(\byear{2022})
\doiurl{10.1126/sciadv.abj4383}
\end{barticle}
\endbibitem

\bibitem[\protect\citeauthoryear{Yin et~al.}{2006}]{BU}
\begin{bchapter}
\bauthor{\bsnm{Yin}, \binits{L.}},
\bauthor{\bsnm{Wei}, \binits{X.}},
\bauthor{\bsnm{Sun}, \binits{Y.}},
\bauthor{\bsnm{Wang}, \binits{J.}},
\bauthor{\bsnm{Rosato}, \binits{M.J.}}:
\bctitle{A 3d facial expression database for facial behavior research}.
In: \bbtitle{7th International Conference on Automatic Face and Gesture
  Recognition (FGR06)},
pp. \bfpage{211}--\blpage{216}
(\byear{2006}).
\bcomment{IEEE}
\end{bchapter}
\endbibitem

\bibitem[\protect\citeauthoryear{Lyons et~al.}{1998}]{JAFFE}
\begin{bchapter}
\bauthor{\bsnm{Lyons}, \binits{M.J.}},
\bauthor{\bsnm{Akamatsu}, \binits{S.}},
\bauthor{\bsnm{Kamachi}, \binits{M.}},
\bauthor{\bsnm{Gyoba}, \binits{J.}},
\bauthor{\bsnm{Budynek}, \binits{J.}}:
\bctitle{The japanese female facial expression (jaffe) database}.
In: \bbtitle{Third International Conference on Automatic Face and Gesture
  Recognition},
pp. \bfpage{14}--\blpage{16}
(\byear{1998})
\end{bchapter}
\endbibitem

\bibitem[\protect\citeauthoryear{Olszanowski et~al.}{2015}]{wsefep}
\begin{barticle}
\bauthor{\bsnm{Olszanowski}, \binits{M.}},
\bauthor{\bsnm{Pochwatko}, \binits{G.}},
\bauthor{\bsnm{Kuklinski}, \binits{K.}},
\bauthor{\bsnm{Scibor-Rylski}, \binits{M.}},
\bauthor{\bsnm{Lewinski}, \binits{P.}},
\bauthor{\bsnm{Ohme}, \binits{R.K.}}:
\batitle{Warsaw set of emotional facial expression pictures: a validation study
  of facial display photographs}.
\bjtitle{Frontiers in psychology}
\bvolume{5},
\bfpage{1516}
(\byear{2015})
\end{barticle}
\endbibitem

\bibitem[\protect\citeauthoryear{Ramis et~al.}{2022}]{SilNet2022}
\begin{barticle}
\bauthor{\bsnm{Ramis}, \binits{S.}},
\bauthor{\bsnm{Buades}, \binits{J.M.}},
\bauthor{\bsnm{Perales}, \binits{F.J.}},
\bauthor{\bsnm{Manresa-Yee}, \binits{C.}}:
\batitle{A novel approach to cross dataset studies in facial expression
  recognition}.
\bjtitle{Multimedia Tools Appl.}
\bvolume{81}(\bissue{27}),
\bfpage{39507}--\blpage{39544}
(\byear{2022})
\doiurl{10.1007/s11042-022-13117-2}
\end{barticle}
\endbibitem

\bibitem[\protect\citeauthoryear{Lisani et~al.}{2017}]{Lisani2017}
\begin{barticle}
\bauthor{\bsnm{Lisani}, \binits{J.-L.}},
\bauthor{\bsnm{Ramis}, \binits{S.}},
\bauthor{\bsnm{Perales}, \binits{F.J.}}:
\batitle{A contrario detection of faces: A case example}.
\bjtitle{SIAM Journal on Imaging Sciences}
\bvolume{10}(\bissue{4}),
\bfpage{2091}--\blpage{2118}
(\byear{2017})
\doiurl{10.1137/17M1118774}
\end{barticle}
\endbibitem

\bibitem[\protect\citeauthoryear{Kazemi and Sullivan}{2014}]{kazemi2014one}
\begin{bchapter}
\bauthor{\bsnm{Kazemi}, \binits{V.}},
\bauthor{\bsnm{Sullivan}, \binits{J.}}:
\bctitle{One millisecond face alignment with an ensemble of regression trees}.
In: \bbtitle{2014 IEEE Conference on Computer Vision and Pattern Recognition},
pp. \bfpage{1867}--\blpage{1874}
(\byear{2014}).
\doiurl{10.1109/CVPR.2014.241}
\end{bchapter}
\endbibitem

\bibitem[\protect\citeauthoryear{Wu and Ji}{2019}]{Wu2019}
\begin{barticle}
\bauthor{\bsnm{Wu}, \binits{Y.}},
\bauthor{\bsnm{Ji}, \binits{Q.}}:
\batitle{{Facial Landmark Detection: A Literature Survey}}.
\bjtitle{International Journal of Computer Vision}
\bvolume{127}(\bissue{2}),
\bfpage{115}--\blpage{142}
(\byear{2019})
\doiurl{10.1007/s11263-018-1097-z}
\end{barticle}
\endbibitem

\bibitem[\protect\citeauthoryear{McReynolds and
  Blythe}{2005}]{MCREYNOLDS200535}
\begin{bchapter}
\bauthor{\bsnm{McReynolds}, \binits{T.}},
\bauthor{\bsnm{Blythe}, \binits{D.}}:
\bctitle{Chapter 3 - color, shading, and lighting}.
In: \beditor{\bsnm{McReynolds}, \binits{T.}},
\beditor{\bsnm{Blythe}, \binits{D.}} (eds.)
\bbtitle{Advanced Graphics Programming Using OpenGL}.
\bsertitle{The Morgan Kaufmann Series in Computer Graphics},
pp. \bfpage{35}--\blpage{56}.
\bpublisher{Morgan Kaufmann},
\blocation{San Francisco}
(\byear{2005}).
\doiurl{10.1016/B978-155860659-3.50005-6}
\end{bchapter}
\endbibitem

\bibitem[\protect\citeauthoryear{Krizhevsky et~al.}{2012}]{Alexnet2012}
\begin{bchapter}
\bauthor{\bsnm{Krizhevsky}, \binits{A.}},
\bauthor{\bsnm{Sutskever}, \binits{I.}},
\bauthor{\bsnm{Hinton}, \binits{G.E.}}:
\bctitle{Imagenet classification with deep convolutional neural networks}.
In: \beditor{\bsnm{Pereira}, \binits{F.}},
\beditor{\bsnm{Burges}, \binits{C.J.}},
\beditor{\bsnm{Bottou}, \binits{L.}},
\beditor{\bsnm{Weinberger}, \binits{K.Q.}} (eds.)
\bbtitle{Advances in Neural Information Processing Systems},
vol. \bseriesno{25}.
\bpublisher{Curran Associates, Inc.}, \blocation{???}
(\byear{2012})
\end{bchapter}
\endbibitem

\bibitem[\protect\citeauthoryear{Simonyan and
  Zisserman}{2015}]{simonyan2015deep}
\begin{botherref}
\oauthor{\bsnm{Simonyan}, \binits{K.}},
\oauthor{\bsnm{Zisserman}, \binits{A.}}:
Very Deep Convolutional Networks for Large-Scale Image Recognition
(2015)
\end{botherref}
\endbibitem

\bibitem[\protect\citeauthoryear{He et~al.}{2015}]{ResNet2015}
\begin{botherref}
\oauthor{\bsnm{He}, \binits{K.}},
\oauthor{\bsnm{Zhang}, \binits{X.}},
\oauthor{\bsnm{Ren}, \binits{S.}},
\oauthor{\bsnm{Sun}, \binits{J.}}:
Deep Residual Learning for Image Recognition
(2015)
\end{botherref}
\endbibitem

\bibitem[\protect\citeauthoryear{Szegedy et~al.}{2015}]{Inception2015}
\begin{botherref}
\oauthor{\bsnm{Szegedy}, \binits{C.}},
\oauthor{\bsnm{Vanhoucke}, \binits{V.}},
\oauthor{\bsnm{Ioffe}, \binits{S.}},
\oauthor{\bsnm{Shlens}, \binits{J.}},
\oauthor{\bsnm{Wojna}, \binits{Z.}}:
Rethinking the Inception Architecture for Computer Vision
(2015)
\end{botherref}
\endbibitem

\bibitem[\protect\citeauthoryear{Chollet}{2017}]{Xception2017}
\begin{bchapter}
\bauthor{\bsnm{Chollet}, \binits{F.}}:
\bctitle{Xception: Deep learning with depthwise separable convolutions}.
In: \bbtitle{Proceedings of the IEEE Conference on Computer Vision and Pattern
  Recognition (CVPR)}
(\byear{2017})
\end{bchapter}
\endbibitem

\bibitem[\protect\citeauthoryear{Howard et~al.}{2019}]{MobileNetV3}
\begin{botherref}
\oauthor{\bsnm{Howard}, \binits{A.}},
\oauthor{\bsnm{Sandler}, \binits{M.}},
\oauthor{\bsnm{Chu}, \binits{G.}},
\oauthor{\bsnm{Chen}, \binits{L.-C.}},
\oauthor{\bsnm{Chen}, \binits{B.}},
\oauthor{\bsnm{Tan}, \binits{M.}},
\oauthor{\bsnm{Wang}, \binits{W.}},
\oauthor{\bsnm{Zhu}, \binits{Y.}},
\oauthor{\bsnm{Pang}, \binits{R.}},
\oauthor{\bsnm{Vasudevan}, \binits{V.}},
\oauthor{\bsnm{Le}, \binits{Q.V.}},
\oauthor{\bsnm{Adam}, \binits{H.}}:
Searching for MobileNetV3
(2019)
\end{botherref}
\endbibitem

\bibitem[\protect\citeauthoryear{Tan and Le}{2021}]{EfficientNetV2}
\begin{botherref}
\oauthor{\bsnm{Tan}, \binits{M.}},
\oauthor{\bsnm{Le}, \binits{Q.V.}}:
EfficientNetV2: Smaller Models and Faster Training
(2021)
\end{botherref}
\endbibitem

\bibitem[\protect\citeauthoryear{Song et~al.}{2014}]{Song2014}
\begin{bchapter}
\bauthor{\bsnm{Song}, \binits{I.}},
\bauthor{\bsnm{Kim}, \binits{H.-J.}},
\bauthor{\bsnm{Jeon}, \binits{P.B.}}:
\bctitle{Deep learning for real-time robust facial expression recognition on a
  smartphone}.
In: \bbtitle{2014 IEEE International Conference on Consumer Electronics
  (ICCE)},
pp. \bfpage{564}--\blpage{567}
(\byear{2014}).
\doiurl{10.1109/ICCE.2014.6776135}
\end{bchapter}
\endbibitem

\bibitem[\protect\citeauthoryear{Li et~al.}{2015}]{WeiNet2015}
\begin{bchapter}
\bauthor{\bsnm{Li}, \binits{W.}},
\bauthor{\bsnm{Li}, \binits{M.}},
\bauthor{\bsnm{Su}, \binits{Z.}},
\bauthor{\bsnm{Zhu}, \binits{Z.}}:
\bctitle{A deep-learning approach to facial expression recognition with candid
  images}.
In: \bbtitle{2015 14th IAPR International Conference on Machine Vision
  Applications (MVA)},
pp. \bfpage{279}--\blpage{282}
(\byear{2015}).
\bcomment{IEEE}
\end{bchapter}
\endbibitem

\bibitem[\protect\citeauthoryear{He et~al.}{2016}]{he2016deep}
\begin{bchapter}
\bauthor{\bsnm{He}, \binits{K.}},
\bauthor{\bsnm{Zhang}, \binits{X.}},
\bauthor{\bsnm{Ren}, \binits{S.}},
\bauthor{\bsnm{Sun}, \binits{J.}}:
\bctitle{Deep residual learning for image recognition}.
In: \bbtitle{2016 IEEE Conference on Computer Vision and Pattern Recognition
  (CVPR)},
pp. \bfpage{770}--\blpage{778}
(\byear{2016}).
\doiurl{10.1109/CVPR.2016.90}
\end{bchapter}
\endbibitem

\bibitem[\protect\citeauthoryear{Szegedy et~al.}{2014}]{szegedy2014going}
\begin{botherref}
\oauthor{\bsnm{Szegedy}, \binits{C.}},
\oauthor{\bsnm{Liu}, \binits{W.}},
\oauthor{\bsnm{Jia}, \binits{Y.}},
\oauthor{\bsnm{Sermanet}, \binits{P.}},
\oauthor{\bsnm{Reed}, \binits{S.}},
\oauthor{\bsnm{Anguelov}, \binits{D.}},
\oauthor{\bsnm{Erhan}, \binits{D.}},
\oauthor{\bsnm{Vanhoucke}, \binits{V.}},
\oauthor{\bsnm{Rabinovich}, \binits{A.}}:
Going Deeper with Convolutions
(2014)
\end{botherref}
\endbibitem

\bibitem[\protect\citeauthoryear{Howard et~al.}{2017}]{howard2017mobilenets}
\begin{botherref}
\oauthor{\bsnm{Howard}, \binits{A.G.}},
\oauthor{\bsnm{Zhu}, \binits{M.}},
\oauthor{\bsnm{Chen}, \binits{B.}},
\oauthor{\bsnm{Kalenichenko}, \binits{D.}},
\oauthor{\bsnm{Wang}, \binits{W.}},
\oauthor{\bsnm{Weyand}, \binits{T.}},
\oauthor{\bsnm{Andreetto}, \binits{M.}},
\oauthor{\bsnm{Adam}, \binits{H.}}:
MobileNets: Efficient Convolutional Neural Networks for Mobile Vision
  Applications
(2017)
\end{botherref}
\endbibitem

\bibitem[\protect\citeauthoryear{Tan and Le}{2020}]{tan2020efficientnet}
\begin{botherref}
\oauthor{\bsnm{Tan}, \binits{M.}},
\oauthor{\bsnm{Le}, \binits{Q.V.}}:
EfficientNet: Rethinking Model Scaling for Convolutional Neural Networks
(2020)
\end{botherref}
\endbibitem

\bibitem[\protect\citeauthoryear{{van der Velden}
  et~al.}{2022}]{VANDERVELDEN2022102470}
\begin{barticle}
\bauthor{\bsnm{{van der Velden}}, \binits{B.H.M.}},
\bauthor{\bsnm{Kuijf}, \binits{H.J.}},
\bauthor{\bsnm{Gilhuijs}, \binits{K.G.A.}},
\bauthor{\bsnm{Viergever}, \binits{M.A.}}:
\batitle{Explainable artificial intelligence (xai) in deep learning-based
  medical image analysis}.
\bjtitle{Medical Image Analysis}
\bvolume{79},
\bfpage{102470}
(\byear{2022})
\doiurl{10.1016/j.media.2022.102470}
\end{barticle}
\endbibitem

\bibitem[\protect\citeauthoryear{Ribeiro et~al.}{2016}]{Ribeiro2016}
\begin{barticle}
\bauthor{\bsnm{Ribeiro}, \binits{M.T.}},
\bauthor{\bsnm{Singh}, \binits{S.}},
\bauthor{\bsnm{Guestrin}, \binits{C.}}:
\batitle{{"Why should i trust you?" Explaining the predictions of any
  classifier}}.
\bjtitle{Proceedings of the ACM SIGKDD International Conference on Knowledge
  Discovery and Data Mining}
\bvolume{13-17-Augu},
\bfpage{1135}--\blpage{1144}
(\byear{2016})
\doiurl{10.1145/2939672.2939778}
\end{barticle}
\endbibitem

\bibitem[\protect\citeauthoryear{Alicioglu and Sun}{2022}]{ALICIOGLU2022502}
\begin{barticle}
\bauthor{\bsnm{Alicioglu}, \binits{G.}},
\bauthor{\bsnm{Sun}, \binits{B.}}:
\batitle{A survey of visual analytics for explainable artificial intelligence
  methods}.
\bjtitle{Computers \& Graphics}
\bvolume{102},
\bfpage{502}--\blpage{520}
(\byear{2022})
\doiurl{10.1016/j.cag.2021.09.002}
\end{barticle}
\endbibitem

\bibitem[\protect\citeauthoryear{Achanta et~al.}{2012}]{SLIC}
\begin{barticle}
\bauthor{\bsnm{Achanta}, \binits{R.}},
\bauthor{\bsnm{Shaji}, \binits{A.}},
\bauthor{\bsnm{Smith}, \binits{K.}},
\bauthor{\bsnm{Lucchi}, \binits{A.}},
\bauthor{\bsnm{Fua}, \binits{P.}},
\bauthor{\bsnm{Süsstrunk}, \binits{S.}}:
\batitle{Slic superpixels compared to state-of-the-art superpixel methods}.
\bjtitle{IEEE Transactions on Pattern Analysis and Machine Intelligence}
\bvolume{34}(\bissue{11}),
\bfpage{2274}--\blpage{2282}
(\byear{2012})
\doiurl{10.1109/TPAMI.2012.120}
\end{barticle}
\endbibitem

\bibitem[\protect\citeauthoryear{Perveen and Mohan}{2020}]{Perveen2020}
\begin{bchapter}
\bauthor{\bsnm{Perveen}, \binits{N.}},
\bauthor{\bsnm{Mohan}, \binits{C.}}:
\bctitle{{Configural Representation of Facial Action Units for Spontaneous
  Facial Expression Recognition in the Wild}}.
In: \bbtitle{15th International Conference on Computer Vision Theory and
  Applications}
(\byear{2020}).
\doiurl{10.5220/0009099700930102}
\end{bchapter}
\endbibitem

\bibitem[\protect\citeauthoryear{Otsu}{1979}]{otsu}
\begin{barticle}
\bauthor{\bsnm{Otsu}, \binits{N.}}:
\batitle{A threshold selection method from gray-level histograms}.
\bjtitle{IEEE Transactions on Systems, Man, and Cybernetics}
\bvolume{9}(\bissue{1}),
\bfpage{62}--\blpage{66}
(\byear{1979})
\doiurl{10.1109/TSMC.1979.4310076}
\end{barticle}
\endbibitem

\bibitem[\protect\citeauthoryear{Manresa-Yee et~al.}{2024}]{manresa2024impact}
\begin{bchapter}
\bauthor{\bsnm{Manresa-Yee}, \binits{C.}},
\bauthor{\bsnm{Ramis}, \binits{S.}},
\bauthor{\bsnm{Gaya-Morey}, \binits{F.X.}},
\bauthor{\bsnm{Buades}, \binits{J.M.}}:
\bctitle{Impact of explanations for trustworthy and transparent artificial
  intelligence}.
In: \bbtitle{Proceedings of the XXIII International Conference on Human
  Computer Interaction}.
\bsertitle{Interacci\'{o}n '23}.
\bpublisher{Association for Computing Machinery},
\blocation{New York, NY, USA}
(\byear{2024}).
\doiurl{10.1145/3612783.3612798}
\end{bchapter}
\endbibitem

\bibitem[\protect\citeauthoryear{Ekman}{1971}]{Ekman1971}
\begin{barticle}
\bauthor{\bsnm{Ekman}, \binits{P.}}:
\batitle{{Universals and cultural differences in facial expressions of
  emotion}}.
\bjtitle{Nebraska Symposium on Motivation}
\bvolume{19},
\bfpage{207}--\blpage{283}
(\byear{1971})
\end{barticle}
\endbibitem

\bibitem[\protect\citeauthoryear{Peterson et~al.}{2018}]{Peterson}
\begin{barticle}
\bauthor{\bsnm{Peterson}, \binits{J.C.}},
\bauthor{\bsnm{Abbott}, \binits{J.T.}},
\bauthor{\bsnm{Griffiths}, \binits{T.L.}}:
\batitle{Evaluating (and improving) the correspondence between deep neural
  networks and human representations}.
\bjtitle{Cognitive Science}
\bvolume{42}(\bissue{8}),
\bfpage{2648}--\blpage{2669}
(\byear{2018})
\doiurl{10.1111/cogs.12670}
{\href{https://arxiv.org/abs/https://onlinelibrary.wiley.com/doi/pdf/10.1111/cogs.12670}{{https://onlinelibrary.wiley.com/doi/pdf/10.1111/cogs.12670}}}
\end{barticle}
\endbibitem

\bibitem[\protect\citeauthoryear{Muttenthaler et~al.}{2023}]{Muttenthaler}
\begin{bchapter}
\bauthor{\bsnm{Muttenthaler}, \binits{L.}},
\bauthor{\bsnm{Linhardt}, \binits{L.}},
\bauthor{\bsnm{Dippel}, \binits{J.}},
\bauthor{\bsnm{Vandermeulen}, \binits{R.A.}},
\bauthor{\bsnm{Hermann}, \binits{K.}},
\bauthor{\bsnm{Lampinen}, \binits{A.}},
\bauthor{\bsnm{Kornblith}, \binits{S.}}:
\bctitle{Improving neural network representations using human similarity
  judgments}.
In: \beditor{\bsnm{Oh}, \binits{A.}},
\beditor{\bsnm{Naumann}, \binits{T.}},
\beditor{\bsnm{Globerson}, \binits{A.}},
\beditor{\bsnm{Saenko}, \binits{K.}},
\beditor{\bsnm{Hardt}, \binits{M.}},
\beditor{\bsnm{Levine}, \binits{S.}} (eds.)
\bbtitle{Advances in Neural Information Processing Systems},
vol. \bseriesno{36},
pp. \bfpage{50978}--\blpage{51007}.
\bpublisher{Curran Associates, Inc.}, \blocation{???}
(\byear{2023}).
\burl{https://proceedings.neurips.cc/paper_files/paper/2023/file/9febda1c8344cc5f2d51713964864e93-Paper-Conference.pdf}
\end{bchapter}
\endbibitem

\bibitem[\protect\citeauthoryear{Geirhos et~al.}{2020}]{Geirhos}
\begin{bchapter}
\bauthor{\bsnm{Geirhos}, \binits{R.}},
\bauthor{\bsnm{Meding}, \binits{K.}},
\bauthor{\bsnm{Wichmann}, \binits{F.A.}}:
\bctitle{Beyond accuracy: quantifying trial-by-trial behaviour of cnns and
  humans by measuring error consistency}.
In: \bbtitle{Proceedings of the 34th International Conference on Neural
  Information Processing Systems}.
\bsertitle{NIPS '20}.
\bpublisher{Curran Associates Inc.},
\blocation{Red Hook, NY, USA}
(\byear{2020})
\end{bchapter}
\endbibitem

\bibitem[\protect\citeauthoryear{Guidotti et~al.}{2018}]{guidotti2018survey}
\begin{botherref}
\oauthor{\bsnm{Guidotti}, \binits{R.}},
\oauthor{\bsnm{Monreale}, \binits{A.}},
\oauthor{\bsnm{Ruggieri}, \binits{S.}},
\oauthor{\bsnm{Turini}, \binits{F.}},
\oauthor{\bsnm{Pedreschi}, \binits{D.}},
\oauthor{\bsnm{Giannotti}, \binits{F.}}:
A Survey Of Methods For Explaining Black Box Models
(2018)
\end{botherref}
\endbibitem

\end{thebibliography}

\end{document}